\documentclass[]{beingbeyond}
\usepackage{enumitem}
\usepackage[toc,page,header]{appendix}

\usepackage[utf8]{inputenc} 
\usepackage[T1]{fontenc}    
\usepackage{hyperref}       
\usepackage{url}            
\usepackage{array}          
\usepackage{booktabs}       
\usepackage{amsfonts}       
\usepackage{nicefrac}       
\usepackage{microtype}      
\usepackage{xcolor}         
\usepackage{xspace}
\usepackage{bm}
\usepackage{bbm}
\usepackage{bbding}
\usepackage{tabularx}
\usepackage{textcomp}
\usepackage{amssymb}
\usepackage{enumitem}
\usepackage{amsmath}
\usepackage{mathtools}
\usepackage{amsthm}
\usepackage{multirow}
\usepackage{makecell}
\usepackage{color}
\usepackage{colortbl}
\usepackage{adjustbox}
\usepackage{caption}
\usepackage{graphicx}
\usepackage{wrapfig}
\usepackage{array}
\usepackage{multicol}
\usepackage{algorithm}
\usepackage{algorithmic}
\usepackage{diagbox}

\definecolor{myyellow}{RGB}{255,192,0}
\definecolor{mygreen}{RGB}{107,170,64}
\definecolor{mywrite}{RGB}{255,227,132}

\title{Universal Dexterous Functional Grasping via Demonstration-Editing Reinforcement Learning}

\author{{\bfseries 
Chuan Mao$^{1}$ \quad
Haoqi Yuan$^{1,2}$ \quad 
Ziye Huang$^{1,2}$ \\
Chaoyi Xu$^{2}$ \quad
Kai Ma$^{1}$ \quad
Zongqing Lu$^{1,2,\dagger}$
}}

\affiliation{{$^{1}$Peking University \quad $^{2}$BeingBeyond}}

\webpage{\url{https://beingbeyond.github.io/DemoFunGrasp/}}
\firstfig[width=1\textwidth, trim={-1cm, 4cm, 6cm, 0cm}, clip][1\textwidth]{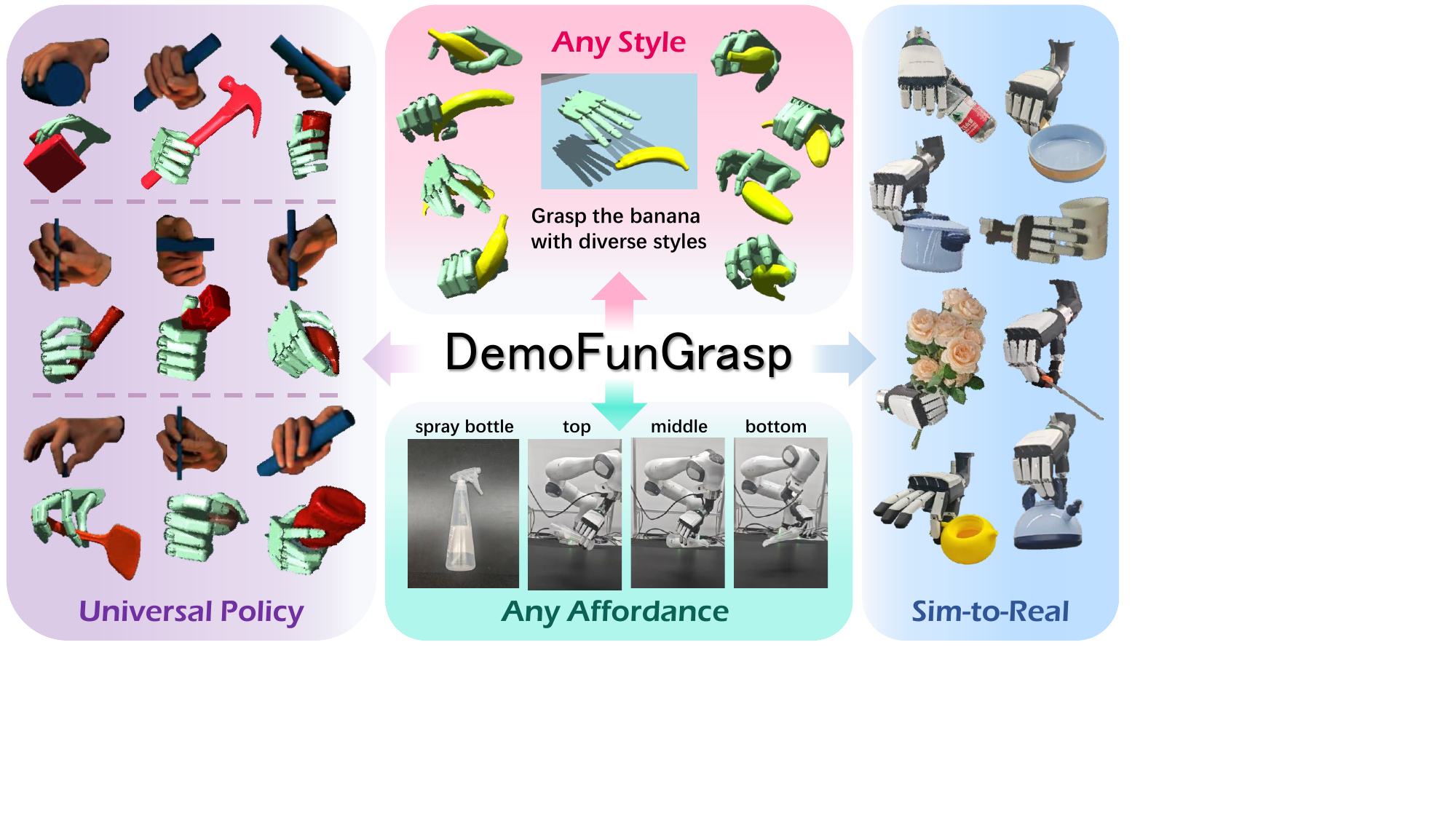}
{\textbf{DemoFunGrasp} is a reinforcement learning framework for universal dexterous functional grasping. The learned policy generalizes to unseen combinations of objects and functional grasping conditions, and achieves zero-shot sim-to-real transfer. For the same object, the policy can produce diverse grasps by adjusting the grasping style and affordance. }
{fig:overview}

\abstract{
Reinforcement learning (RL) has achieved great success in dexterous grasping, significantly improving grasp performance and generalization from simulation to the real world. However, fine-grained functional grasping, which is essential for downstream manipulation tasks, remains underexplored and faces several challenges: the complexity of specifying goals and reward functions for functional grasps across diverse objects, the difficulty of multi-task RL exploration, and the challenge of sim-to-real transfer. 
In this work, we propose \textbf{DemoFunGrasp} for universal dexterous functional grasping. We factorize functional grasping conditions into two complementary components — grasping style and affordance — and integrate them into an RL framework that can learn to grasp any object with any functional grasping condition. To address the multi-task optimization challenge, we leverage a single grasping demonstration and reformulate the RL problem as one-step demonstration editing, substantially enhancing sample efficiency and performance.
Experimental results in both simulation and the real world show that DemoFunGrasp generalizes to unseen combinations of objects, affordances, and grasping styles, outperforming baselines in both success rate and functional grasping accuracy. In addition to strong sim-to-real capability, by incorporating a vision-language model (VLM) for planning, our system achieves autonomous instruction-following grasp execution. 
}

\checkdata[Date]{December 2, 2025}

\definecolor{BlockC}{gray}{0.98}  
\definecolor{BlockA}{RGB}{191,211,230}
\definecolor{BlockB}{RGB}{199,233,192}

\begingroup
\footnotetext[2]{Correspondence to Zongqing Lu $<$lu@beingbeyond.com$>$.}
\endgroup

\begin{document}

\maketitle

\section{Introduction}
\label{sec:intro}


Manipulation is a core capability of embodied intelligence, with grasping serving as its indispensable foundation. 
An effective grasp not only provides stability but also enables subsequent tool-use and manipulation tasks. 
Consequently, many studies~\cite{xu2023unidexgrasp,wan2023unidexgrasp++,huang2024efficient,zhang2025robustdexgrasp,zhong2025dexgraspvla} have explored learning robust grasping policies to achieve adaptive, closed-loop grasping in real-world tabletop settings. 
However, most existing approaches focus solely on achieving mechanical stability and do not consider the functional requirements of the grasp. 

Functional grasping naturally highlights the advantages of dexterous robotic hands over parallel grippers. 
With a higher number of degrees of freedom (DoFs), dexterous hands can adapt to diverse object geometries and support a broad range of grasping styles. 
Recent studies have explored synthesizing human-like grasping poses by generating datasets~\cite{wei2025afforddexgrasp,liu2022hoi4d,brahmbhatt2019contactdb, chen2025dexonomy} or by learning from large-scale human data~\cite{dexvlg25,bahl2023affordances,huang2025fungrasp}. 
However, these methods typically rely heavily on human supervision, and the limited availability of high-quality data restricts their generalization to unseen objects. 
Furthermore, the generated poses are usually executed through open-loop planning, which limits their practicality in real-world tabletop scenarios.
Other works~\cite{wu2024dexterous,agarwal2023dexterous} leverage reinforcement learning (RL) in large-scale simulation to obtain closed-loop functional grasping policies. 
However, the high dimensionality of a dexterous hand's action space, together with the multi-task optimization challenge introduced by diverse objects and functional grasping styles, creates substantial difficulties for RL algorithm design and significantly limits the performance.

To tackle these challenges, we propose \textbf{DemoFunGrasp}, an effective RL framework that can learn to grasp \textit{any object} under \textit{any functional grasping condition}. 
We first decompose each functional grasping condition into two components: \textit{affordance} and \textit{grasping style}. The affordance specifies the region of the object to grasp (where to grasp), and the grasping style specifies the reference hand pose (how to grasp). They provide a complete description of the intended functional grasp.
To learn a universal policy across all objects and functional conditions, we introduce an RL framework that incorporates these conditions into both the policy observations and the reward function. Diverse objects, affordances, and grasping styles are sampled in parallel simulation to train a universal policy.
To address the multi-step and multi-task exploration challenge of this RL problem, we build on recent advances in demonstration-editing RL~\cite{yuan2025demograsp}. We collect one grasping demonstration and train a policy that outputs a wrist transformation (determining where to grasp) and a hand-style adaptation delta action (determining how to grasp). These actions edit the object-centric robot actions in the demonstration, which is then replayed for trial-and-error learning. This formulation reduces the problem to a single-step RL task and tightly links the action space with the functional grasping conditions, leading to significantly improved sample efficiency.

Experiments on 3{,}200 DexGraspNet objects~\cite{wang2022dexgraspnet} show that DemoFunGrasp achieves state-of-the-art performance, with higher affordance accuracy and greater style diversity compared with prior work. 
We further demonstrate the extensibility of our approach by applying it to multiple dexterous hand embodiments without hyperparameter tuning, achieving an overall success rate above 77\% on human-intended grasping styles and affordances. 
To enable sim-to-real transfer, we collect 30{,}000 successful rollouts in simulation using the RL policy and distill them into an RGB-based policy through imitation learning. 
The resulting vision-based policy transfers to a real robot in a zero-shot manner and achieves a 71\% success rate. 
Finally, by integrating a vision-language model (VLM) with the DemoFunGrasp policy, we construct an autonomous grasping system that reaches an average real-world success rate of 64.4\% for functional grasping guided by language instructions.

Our main contributions are summarized as follows:
\begin{itemize}
    \item We introduce DemoFunGrasp, a framework for universal dexterous functional grasping that can handle a wide range of objects under any affordance and grasping style.
    \item We integrate grasping affordances and styles into the observation space, reward function, and action space of demonstration-editing RL. This provides informative learning signals and significantly improves sample efficiency, allowing the policy to address the challenging multi-task optimization problem in functional grasping.
     \item Compared with prior work, our approach not only achieves higher success rates but also follows diverse grasping styles and affordances more accurately. We demonstrate that the learned policy can act as a low-level executor for VLMs, enabling autonomous language-guided functional grasping.
\end{itemize}


\section{Related Work}
\label{sec:related}

\subsection{Dexterous Functional Grasping}

Unlike stability-oriented grasping, functional grasping seeks to produce grasp configurations that are aligned with an object’s intended use. Early studies constructed human grasp taxonomies and analyzed the mapping between grasp forms and task intent~\cite{9812388,feix2015grasp}, offering foundational guidance for functional grasp synthesis.
More recent approaches integrate human priors and semantic cues to generate diverse functional grasps through optimization~\cite{chen2025dexonomy}, data-driven learning~\cite{wang2022dexgraspnet,grady2021contactopt}, cross-category contact transfer~\cite{wu2024cross}, and demonstration retargeting~\cite{huang2025fungrasp}.
However, these grasp-generation frameworks are typically open-loop and exhibit limited reliability when executed in tabletop manipulation tasks. Recent RL-based methods~\cite{agarwal2023dexterous, wu2024dexterous} have demonstrated that closed-loop functional grasping policies can be learned, but the large multi-task exploration space leads to challenging optimization and limited generalization.
Other approaches~\cite{dexvlg25,wei2024grasp} utilize state-of-the-art VLMs for semantic conditioning, but they depend heavily on large-scale annotations and inherit the well-known robustness limitations of current VLMs. In this work, we propose an efficient RL framework that can learn a universal functional grasping policy for diverse objects and conditions, while enabling reliable sim-to-real transfer.


\subsection{Affordance-Based Grasping}

Affordance-based grasping aims to localize functionally meaningful regions that reflect intended object use. Classical methods~\cite{song2015learning,detry2011learning,kokic2017affordance,ardon2019learning,duan2021robotics} predict affordance maps from visual input, revealing grasp-supporting structures such as handles or rims. 
Recent works extend this paradigm to dexterous hands and multimodal reasoning~\cite{mandikal2021learning,mandikal2022dexvip,wei2025afforddexgrasp,wei2025omnidexgrasp}. 
However, most approaches predominantly emphasize perception and affordance inference, while providing limited consideration for closed-loop control or execution reliability, which restricts their effectiveness in real-world manipulation settings. 
Our work bridges the gap between affordance perception and grasping execution by learning an end-to-end RL policy.

\subsection{RL for Dexterous Manipulation}

RL has shows strong capability for dexterous manipulation in simulation~\cite{rajeswaran2017learning,popov2017data,openai2018learning}, but early methods exhibit limited generalization and weak sim-to-real transfer.  
Subsequent work improves data efficiency and robustness through techniques such as residual learning~\cite{huang2024efficient}, human-in-the-loop adaptation~\cite{luo2025precise}, curriculum learning~\cite{xu2023unidexgrasp,wan2023unidexgrasp++,zhang2025robustdexgrasp}, and cross-embodiment transfer~\cite{yuan2024cross}.  
Demonstration-augmented approaches further reduce learning complexity by leveraging a small number of human demonstrations and formulating residual RL~\cite{zhou2024learning,zhu2025dexflywheel} or one-step demonstration-editing RL~\cite{yuan2025demograsp}, which alleviates the exploration burden. Works on sim-to-real transfer and tactile feedback~\cite{lin2025sim,hu2025dexterous} further enhance real-world reliability in contact-rich scenarios. Building on these advances, our approach integrates demonstration editing RL to address the multi-task optimization challenge, learning a style- and affordance-conditioned closed-loop policy that tightly couples functional semantics with dexterous control.



\section{Method}
\label{sec:Methods}

\begin{figure*}[t!]
  \centering
  \includegraphics[width=1.0\linewidth, trim=15 120 15 5, clip]{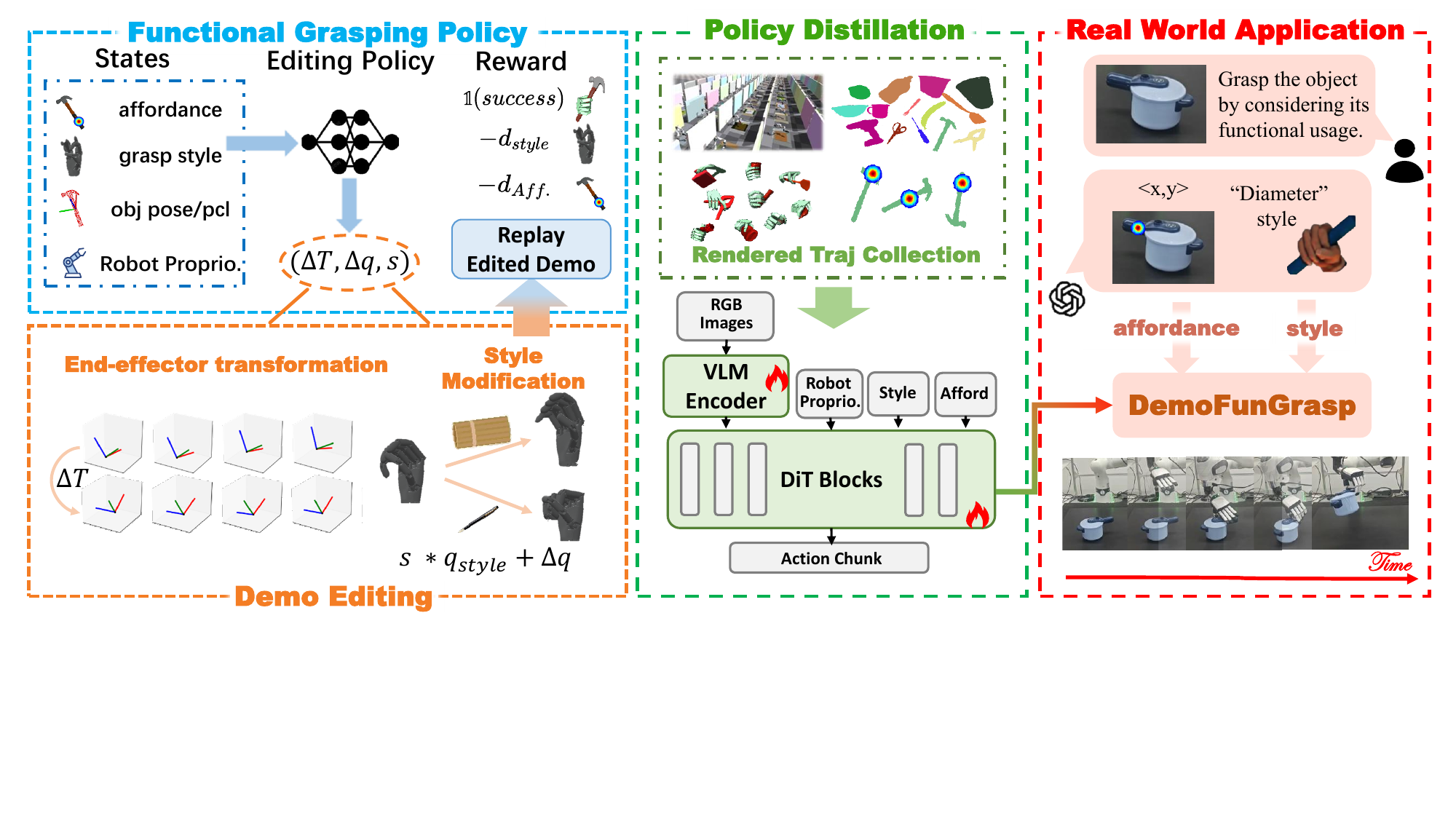}
  \vspace{-1.5em}

\caption{
Overview of \textbf{DemoFunGrasp}. 
(1) \textbf{Demonstration editing:} A source demonstration is adapted through end-effector transformation and object-geometry–aware hand style adjustment. 
(2) \textbf{Functional grasping policy learning:} An affordance- and style-conditioned one-step RL policy is trained. 
(3) \textbf{Vision-based imitation:} The learned policy is transferred to RGB observations for closed-loop, vision-based execution. 
(4) \textbf{Real-world deployment:} The vision-based policy is guided by a Vision-Language Model (VLM) for autonomous planning and execution.
}
  \label{fig:pic2}
  \vspace{-1em}
\end{figure*}

\subsection{Problem Formulation}

We formulate {dexterous functional grasping} as a one-step Markov Decision Process (MDP) following \citet{yuan2025demograsp}. Under the state-based setting, the agent observes 
\[
(\mathbf{s}_r, \mathbf{s}_o, \mathbf{x}_o, p_{\text{afford}}, l_{\text{style}}).
\]
Here, $\mathbf{s}_r$ denotes the robot end-effector 6D pose and $\mathbf{s}_o$ denotes the target object pose. $\mathbf{x}_o$ represents the complete object point cloud. $p_{\text{afford}}\in\mathbb{R}^3$ indicates the affordance point in 3D space, such as a cup handle or a lid edge, representing the desired functional contact region. $l_{\text{style}}$ is a one-hot vector encoding the grasping style category.

The policy outputs an action $a=(\Delta T, \Delta q, k)$, where $\Delta T$ updates the end-effector pose, and $\Delta q$ together with $k$ modulates hand joint scaling coefficient relative to the target grasping style, conditioned on the object geometry.

For vision-based imitation learning, we use the following observation:
\[
(\mathbf{s}_r, \mathbf{r},c_{\text{afford}},l_{\text{style}}),
\]
where $\mathbf{r}= (R, G, B)$ and $c_{\text{afford}}$ denotes the 2D projection of the affordance point in the RGB image. The visual policy predicts the end-effector pose and hand joint angles $(t, r, q)$ in a closed-loop manner, providing smooth control based on raw image input.

The state-based policy, which has access to privileged geometric information, accelerates RL and provides high-quality expert data. The vision-based policy emphasizes perceptual robustness for sim-to-real transfer.


\subsection{One-Step Demonstration-Editing RL}

Unlike conventional multi-step RL, which suffers from unstable optimization and high sample complexity in high-DoF manipulation, our framework simplifies functional grasping into a {one-step RL} problem.  
We observe that grasp success and functional accuracy can be effectively optimized by editing a single high-quality demonstration with residual actions learned through policy updates. This substantially reduces exploration difficulty and training cost while preserving the naturalistic grasp behavior.


\textbf{Demonstration Representation.} Each demonstration $D=\{(\mathbf{p}_t^{\mathrm{ee\text{-}obj}},\mathbf{q}_t^{\mathrm{ref\text{-}hand}})\}_{t=0}^{T^D}$ records the full end-effector–to–object trajectory and the corresponding hand joint sequence. At timestep t, $\mathbf{p}_t$ denotes the end-effector pose and $\mathbf{q}_t$ denotes the hand joint position.
Unlike conventional three-stage static grasp data (pre-, grasp-, and post-contact), we preserve the continuous temporal evolution, including the timing of finger closure and the subtle compliance behaviors during contact formation. This temporal richness allows our one-step editing algorithm to interpolate motion in a physically plausible manner. The interpolation details are provided in Sec.~\ref{subsec:Demonstration_Style_Aware_Editing}.

Given a demonstration, the policy predicts $\{\Delta T, \Delta q, k\}$ as residual corrections to the base trajectory.  
The edited grasp is then executed once, and the reward metric outcome provides the RL signal for policy optimization algorithm. 
In this way, the policy effectively learns a smooth mapping from conditioned inputs to small corrective actions to improve grasp stability and contact precision.  
This “one-step editing” mechanism reformulates RL into a tractable refinement process rather than full motion synthesis, significantly reducing the difficulty of dexterous manipulation learning.

\textbf{Reward Function.}  
The total reward combines functional alignment, style consistency, and contact smoothness:
\[
r = \lambda_{\text{afford}}r_{\text{afford}} + \lambda_{\text{close}}r_{\text{close}} + \lambda_{\text{qpos}}r_{\text{qpos}}+r_{\text{success}}.
\]
$r_{\text{success}}$ rewards successful grasps, $r_{\text{afford}}$ and $r_{\text{close}}$ encourage spatial alignment with the desired affordance point, while $r_{\text{qpos}}$ preserves stylistic consistency. Together, these terms ensure that edited grasps remain human-like and functional while adapting to novel object geometries. Their definition is in the following section.


\subsection{Style-Aware Demonstration Editing}
\label{subsec:Demonstration_Style_Aware_Editing}

\textbf{Styles Selection.}  
We adopt the grasp taxonomy of Feix etal.~\cite{feix2015grasp} as an initial reference, and empirically refine it through large-scale simulation experiments. Our experiments reveal two key observations:  
(1) when normalized for object scale, multiple grasp styles converge to nearly identical hand configurations and contact patterns in successful cases, indicating redundancy;  
(2) for table-top manipulation, some grasping styles are physically infeasible, yielding consistently low success rates due to workspace constraints or unstable wrist orientation.  
Based on these findings, we prune the grasp styles to retain a representative subset, removing redundancies and styles with consistently poor performance during training.

Consequently, we adopt nine representative styles for the Shadow Hand (e.g., palmar pinch, lateral, small diameter), and four for the Inspire Hand, whose lower degrees of freedom naturally constrain its expressive range.  
Each selected style is parameterized by a canonical joint configuration $\mathbf{q}_{\text{pos}}$ and a contact mask that specifies the intended contact fingers.  
Examination of various human functional grasping patterns suggests that a functional grasp is determined once the hand’s style, contact points, and the object's affordance point are specified..

\textbf{Geometry-Conditioned Hand Pose Editing.}  
Because grasp feasibility is highly sensitive to local geometry, we allow residual scaling and joint adjustments $(k, \Delta q)$ to adapt the hand posture according to the sampled object geometry $\mathbf{x}_o$. Our target hand joint configuration is $\mathbf{q}_{\text{pos}}^* = k\cdot \mathbf{q}_{\text{pos}} + \Delta \mathbf{q}$. Although the editing relies only on a single demonstration, this parameterization supports generalization to unseen object shapes via the learned geometric embedding. To balance the success rate and style intention, we design style reward:
\[
r_{\text{qpos}}
=   \exp\!\left(-\|\mathbf{q}_{\text{pos}} - \mathbf{q}_{\text{pos}}^*\|_2\right).
\]

\textbf{Motion Interpolation.}  
Given an edited target configuration $\mathbf{q}_{\text{pos}}^*$, we interpolate the motion along the reference trajectory using a fractional coefficient along the reference trajectory:
\[
\mathbf{q}_i = \mathbf{q}_0^{\mathrm{ref}} + f \cdot (\mathbf{q}_i^{\mathrm{ref}} - \mathbf{q}_0^{\mathrm{ref}}),\quad f = \frac{\mathbf{q}_{\text{pos}}^* - \mathbf{q}_0^{\mathrm{ref}}}{\mathbf{q}_{T_l}^{\mathrm{ref}} - \mathbf{q}_0^{\mathrm{ref}}}
\]
This interpolation produces smooth joint trajectories without requiring collision checks and preserves temporal coherence between fingers and object motion.


\subsection{Affordance Conditions}

\textbf{Sampling Strategy.}  
To provide rich functional supervision, it is necessary to sample diverse affordance points on the objects' surface that are physically feasible for grasping. We estimate an affordance likelihood distribution over the object point cloud using surface-normal alignment and the object’s canonical pose after each environment randomly reset. In each training episode, a single affordance candidate is sampled from this distribution. This stochastic conditioning enables the policy to discover multiple valid functional grasp configurations for the same object.

\textbf{Reward Shaping.}  
We adopt a hierarchical reward consisting of a sparse proximity term $r_{\text{afford}}$ and a dense alignment term $r_{\text{close}}$:
\[
r_{\text{afford}} = \mathbb{I}\left(\text{success}\right)
  \mathbb{I}\!\left(d^{T-1} < \frac{\text{obj}_{\text{bb}}}{\gamma}\right)\exp\left(-d^{T-1}\right),
\]
\[
r_{\text{close}} = \mathbb{I}\left(d^{0:T-1}_{\text{min}} < \text{threshold}\right)
\]

Here, $d^{0:T-1}_{\text{min}}$ denotes the minimum distance between the affordance point and the hand contact point along the trajectory, independent of grasp success. $\text{obj}_{\text{bb}}$ denotes the length of the longest edge of the object's mesh bounding box, and $\gamma$ is a hyperparameter. This object-scale normalization improves affordance accuracy across objects of different sizes.

The hierarchical reward design encourages early-stage exploration toward the affordance region via $r_{\text{close}}$, and fine-grained affordance alignment in later stages via $r_{\text{afford}}$, providing complementary guidance essential to stable and precise functional grasping.


\subsection{Vision-Based Sim-to-Real Transfer}

\textbf{Data Collection.}  
After convergence, we deploy the trained policy to collect large-scale demonstrations by executing successful grasps across a diverse set of object instances. Each sample consists of synchronized robot state observations, RGB images, functional conditions and control actions, forming a dataset for visual imitation learning.

\textbf{Network Architecture.}   
We evaluate three architectural variants to predict continuous action sequences $(t, r, q)$: (i) ACT~\cite{zhao2023learning} with a pretrained ViT encoder, (ii) a ViT-based diffusion policy, and (iii) a diffusion transformer equipped with a Vision-Language Model encoder~\cite{bjorck2025gr00t}.  
Our experiments show that using a VLM encoder with a DiT backbone achieves the best performance, as it effectively better captures the multi-modal structure of grasp strategies and better models the inherent uncertainty across different object instances. This benefit is most pronounced when different grasp styles correspond to distinct affordance semantics.

\textbf{Domain Randomization.}  
To narrow the sim-to-real gap, we perform aggressive domain randomization in IsaacGym, varying object and table textures, lighting directions and intensities, camera extrinsics, and initial object and robot poses.

\textbf{Affordance Projection.}  
Each 3D affordance point is projected onto the image plane using camera intrinsics and extrinsics, providing an explicit spatial cue for the visual encoder.


\section{Experiments}
\label{sec:experiments}
We conduct experiments to answer the following questions:
(1) How does our method perform in terms of grasp success rate, affordance alignment, and grasp-style diversity?
(2) How do the reward design and editing mechanism influence functional grasp performance?
(3) How effectively does our method transfer from simulation to real-world settings?

\subsection{Experimental Settings}

\textbf{Train/Eval Settings.}
We train our policy in IsaacGym\citep{isaacgym} using reinforcement learning. For state-based training, we construct a mixed-object dataset by combining YCB~\cite{Calli_2015} and DexGraspNet~\cite{wang2022dexgraspnet}, yielding a broad distribution of object geometries. During training, object poses are randomized within a \(50 \times 50\,\mathrm{cm}\) square region; affordance points and desired hand styles are selected randomly. We use PPO~\citep{ppo} as the underlying RL algorithm.

Once the state-based policy is obtained, we deploy it in simulation to collect 30k trajectories with randomization. These trajectories serve as supervision for training the vision-based policy. To evaluate the vision-based policy, we select three object categories for simulation and real-world experiments, respectively (Appendix~\ref{sec:Objects_in_the_experiment}).

\textbf{Evaluation Metrics.}\ In the absence of fully comparable baselines for functional grasping, we benchmark our policy against existing grasp-generation and grasp-policy methods using four metrics:
    \begin{enumerate}[leftmargin=!, labelwidth=1.5em, labelsep=0.75em]
        \item \textbf{Grasp Success Rate (GSR)} is the fraction of successful grasps over the evaluation set. A grasp is considered successful if the object is lifted by at least 10 cm and remains stable for 20 time steps.
        \item \textbf{Success Affordance Distance (SAD)} (simulation only) measures how closely the executed grasp aligns with the intended functional region. It is computed as the Euclidean distance between the target affordance point and style-aware contact point at the time of lift.
        \item \textbf{Style Diversity (SD)} (simulation only) quantifies the diversity of successful grasping strategies by computing the average Euclidean distance between all pairs of successful hand joint configurations. 
        \item \textbf{Style Accuracy (SA)} (simulation only) measures the fraction of successful grasps that match the conditioned style:
        \[
        SA = \frac{{\text{style match \& success}}}{{\text{success}}},
        \]
        where a \emph{style match} occurs if the executed grasp adopts the same style as the target condition. Higher SA indicates stronger adherence to the intended grasp style.
        
    \end{enumerate}

\textbf{Real-World Settings.} 
We conduct real-world experiments using a 6-DoF Inspire Hand (with 6 active and 6 passive joints) mounted on a 7-DoF Franka 3 arm. For evaluating vision-based policies, we employ two RealSense D435i cameras positioned diagonally around the table. The complete workspace configuration is shown in Appendix~\ref{sec:The_Robot_workspace}.

\subsection{Affordance-Conditioned Grasping}

Existing works in affordance-based grasping lack a unified metric for execution accuracy. To evaluate DemoFunGrasp's performance across functional regions, we compare the diversity of successful grasp regions and mean success affordance distance with DemoGrasp \citep{yuan2025demograsp}, a state-of-the-art method.

We collect contact data from 1,000 simulated environments and visualize the contact point distributions in Fig.~\ref{fig:contact_points}.  
As shown, DemoGrasp tends to focus on the tightest or most geometrically stable region of the object (e.g., purple points in the right hammer image). In contrast, DemoFunGrasp, conditioned on randomly sampled affordance points, generates diverse contact locations across different functional parts such as handles, rims, or edges, thereby enabling genuinely functional grasps.

We further quantify this effect by computing the mean distance between the achieved contact point and the target affordance point in successful trials, as reported in Table~\ref{tab:affordance_dist}.
Across both object categories seen and unseen in the training set, DemoFunGrasp reduces this affordance distance by over 3\,cm compared to DemoGrasp, demonstrating its stronger alignment with the intended functional regions.

\begin{minipage}{0.64\textwidth}
    \captionof{table}{\textbf{Mean Success Affordance Distance(SAD) (cm) between Achieved Contact Region and Target Affordance Point.}}
    \label{tab:affordance_dist}
    \vspace{-1mm}
    \centering
    \begin{tabular}{lccc}
    \toprule
    \textbf{Model} & \textbf{Train} & \textbf{Seen Cat.} & \textbf{Unseen Cat.} \\
    \midrule
    DemoGrasp & 6.29 & 6.27 & 6.20  \\
    \textbf{ours} & \textbf{3.03} & \textbf{3.02} & \textbf{3.21} \\
    \bottomrule
    \end{tabular}
\end{minipage}%
\hfill
\begin{minipage}{0.35\textwidth}
    \centering
    \includegraphics[width=0.60\linewidth, trim={0.1cm, 0.cm, 0.2cm, 0.1cm}, clip]{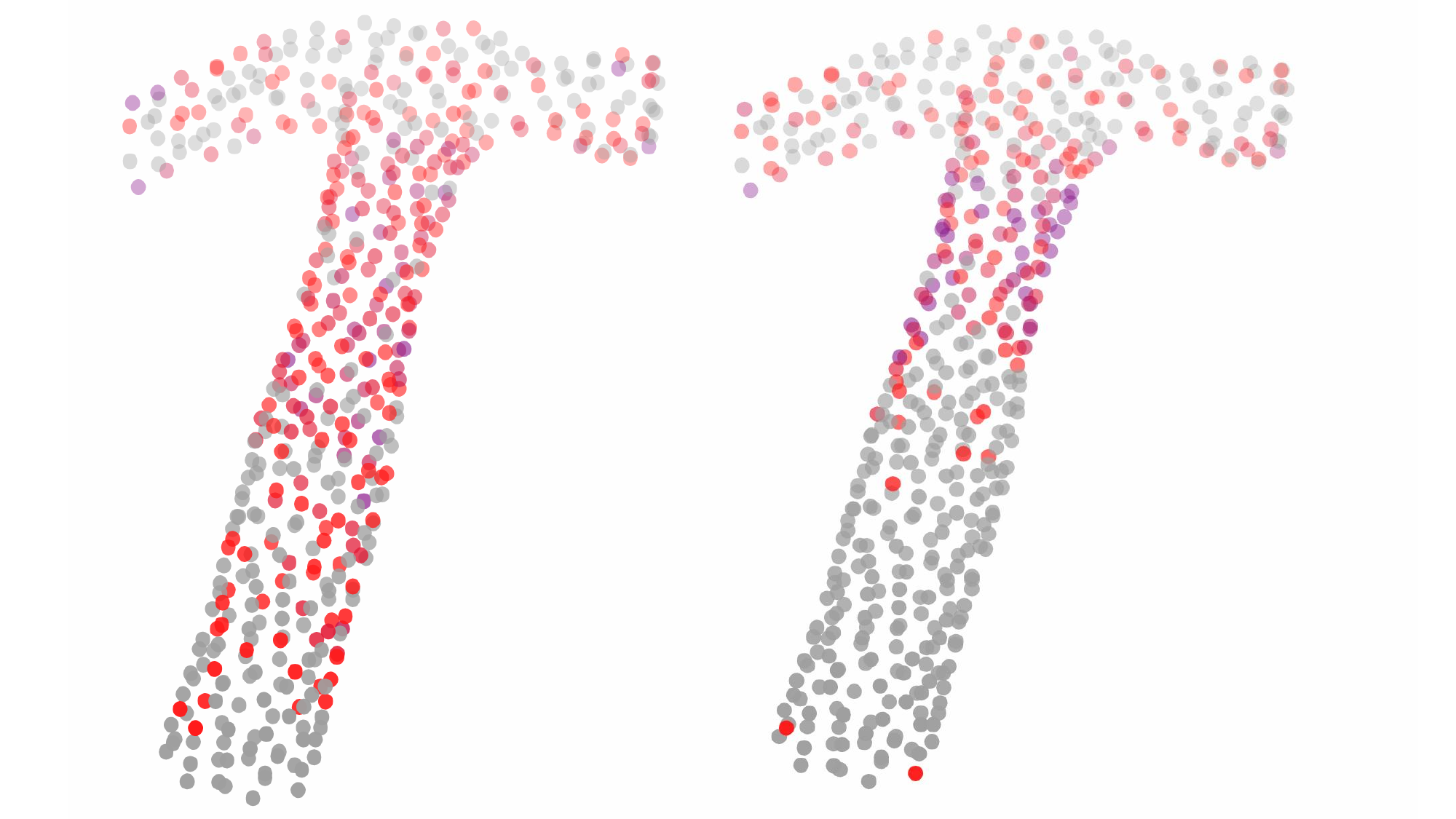}
    \vspace{-4mm}
    \captionof{figure}{\textbf{The contact points distribution. Left: DemoFunGrasp, right: DemoGrasp.}}
    \label{fig:contact_points}
\end{minipage}

\subsection{Diversity of Grasp Styles }
\label{subsec:Diversity_of_Style}

DemoFunGrasp effectively avoids generating hand poses that are ambiguous or physically implausible in real-world scenarios. In contrast, baseline policies trained purely via RL sometimes generate implausible gestures, such as extending the middle finger while contracting the others, which seldom appear in natural human manipulation or functional grasping.

To quantify grasp diversity and functional plausibility, we compare our method against UniDexGrasp~\cite{xu2023unidexgrasp}, a state-of-the-art grasp generation and RL training pipeline.

\begin{minipage}{0.58\textwidth}
    \captionof{table}{\textbf{Grasp Success Rate (GSR) and Style Diversity (SD) compared to UniDexGrasp.}}
    \label{tab:grasp_unidexgrasp}
    \vspace{-1mm}
    \centering
    \begin{tabular}{l|cc|cc}
    \toprule
    \multirow{2}{*}{\textbf{Method}} &
    \multicolumn{2}{c|}{\textbf{Seen Cat.}} &
    \multicolumn{2}{c}{\textbf{Unseen Cat.}} \\
    \cmidrule(r){2-3}\cmidrule(l){4-5}
     & \textbf{GSR}$\uparrow$ & \textbf{SD}$\uparrow$ & \textbf{GSR}$\uparrow$ & \textbf{SD}$\uparrow$  \\
    \midrule
    UniDexGrasp & 74.3 & 1.00 & 70.8 & 1.00 \\
    \textbf{Ours}  & 76.26 & \textbf{1.48} & 71.65 & \textbf{1.44} \\
    \textbf{Ours (best style)}  & \textbf{81.74} & \textbf{1.48} & \textbf{78.68} & 1.36 \\
    \bottomrule
    \end{tabular}
\end{minipage}%
\hfill
\begin{minipage}{0.39\textwidth}
    \centering
    \includegraphics[width=0.95\linewidth, trim={0.1cm, 0.3cm, 0.1cm, 0.3cm}, clip]{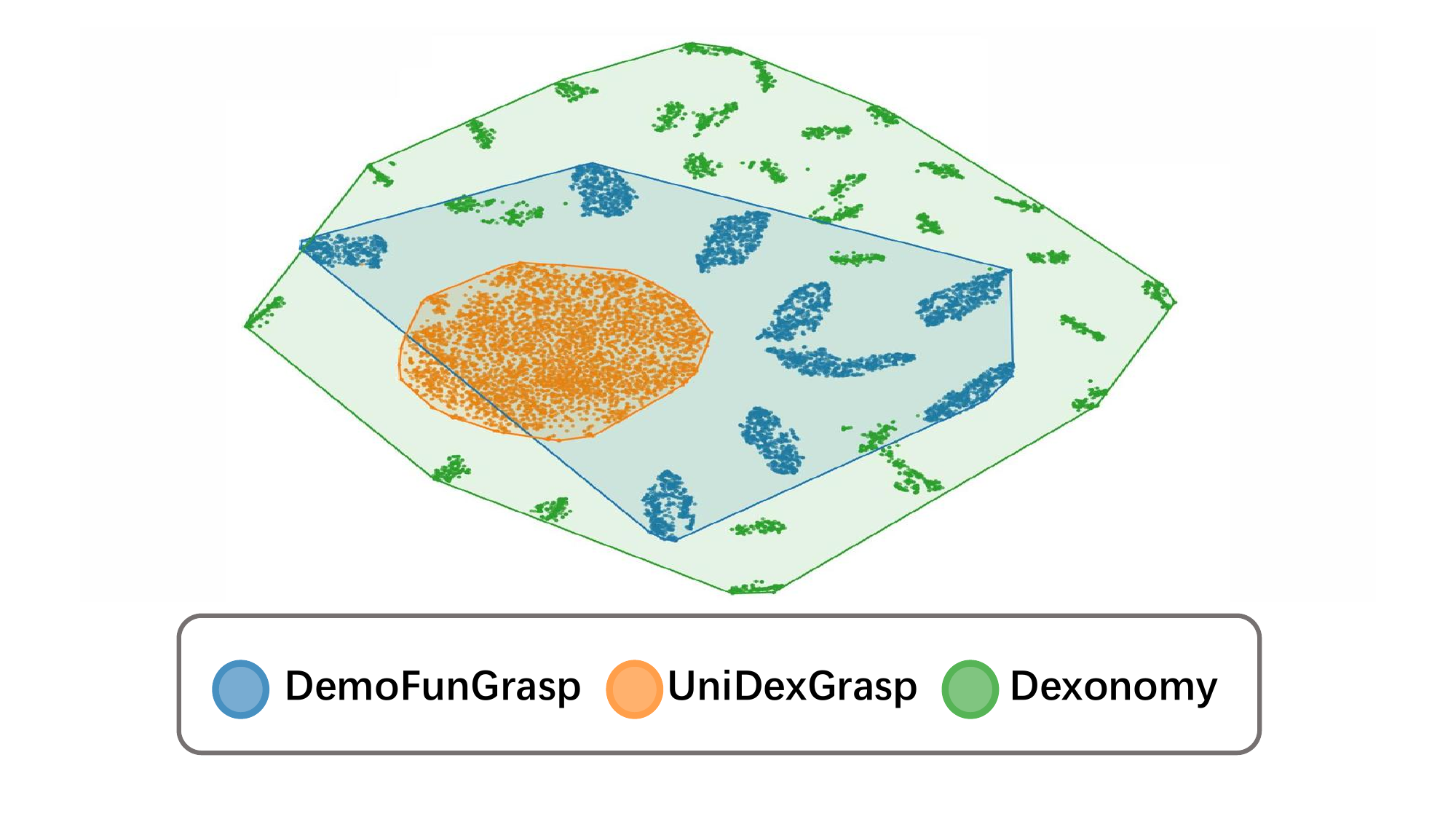}
    \vspace{-4mm}
    \captionof{figure}{\textbf{Grasp style diversity of UniDexGrasp (an RL method), Dexonomy (a grasp synthesis method), and ours.}}
    \label{fig:qpos_tsne}
\end{minipage}

As shown in Table~\ref{tab:grasp_unidexgrasp}, conditioning on style and affordance enables DemoFunGrasp to maintain stable success rates while substantially improving grasp style diversity. During training, the average success rates of DemoFunGrasp and UniDexGrasp are 77.2\% and 79.4\%, respectively. Although our method achieves a slightly lower success rate on the training set, it generalizes better to unseen categories, demonstrating stronger robustness. Because DemoFunGrasp explicitly models \emph{local geometric features}, its grasp success rate remains nearly unchanged across test domains. Moreover, its style diversity reaches almost 1.5$\times$ that of UniDexGrasp, indicating its ability to generate a broader range of physically meaningful and functionally distinct grasps.

We further report performance for the \emph{best-performing style} per object. For each test instance, we replay all style candidates and record the most successful one. The variance across styles stays within 10\%, showing that random sampling of style labels followed by RL refinement yields well-balanced performance among grasp styles. However, selecting only the best style naturally reduces diversity—especially on unseen categories—implying that a compact subset of high-quality styles tends to dominate in robustness.

Many existing grasp-generation methods overlook object–table interactions and often restrict object orientation while permitting arbitrary hand approaches, complicating fair comparison in tabletop settings. To highlight differences between existing paradigms and ours, we visualize hand joint embeddings using t-SNE (Fig.~\ref{fig:qpos_tsne}) and evaluate three settings: the universal grasp model (UniDexGrasp), the functional grasp synthesis framework (Dexonomy), and our method (based on 9 initial styles selected from Dexonomy).

After RL optimization, styles remain well-separated (blue convex hulls), confirming that our conditioning scheme preserves style distinctiveness. A slight cluster shift is also observed, indicating that optimization-based grasp generation tends to adjust configurations toward more stable and contact-consistent postures.


\subsection{Ablation Study}
\label{subsec:Ablation_Study}

We conduct an ablation study on the state-based policy to evaluate the contributions of reward design and style adaptation mechanisms. Table~\ref{tab:Ablation} summarizes the results, showing that each component plays a critical role in the overall performance.

\textbf{Affordance Reward.}\ Removing the affordance reward slightly increases the grasp success rate but substantially increases the success affordance distance, indicating that the policy tends to grasp at positions that are less functionally meaningful. This validates that the affordance reward effectively guides the policy toward target functional regions rather than merely maximizing stability.

\textbf{Object Size Clipping.} When the clipping threshold does not account for object size, large objects tend to yield lower affordance rewards, resulting in suboptimal contact accuracy. Introducing object size clipping mitigates the training instability caused by the wide variation in object sizes. As shown in Fig.~\ref{fig:afford_dist_distribution}, the affordance distance distribution indicates that the size normalization parameter $k$ effectively functions as a clipping factor, alleviating performance bottlenecks in large-object affordance alignment.

\textbf{Style Disturbance.}\ Eliminating style disturbance severely harms performance reducing GSR to 58.67\%. This highlights the importance of style adaptation: perturbation-based exploration ensures both robust and diverse grasping behaviors.

\begin{minipage}{0.62\textwidth}
    \captionof{table}{\textbf{Comprehensive Ablation Study on Model Components.}}
    \label{tab:Ablation}
    \vspace{-1mm}
    \centering
    \begin{tabular}{lccc}
    \toprule
    \textbf{Method} & \textbf{GSR}$\uparrow$ & \textbf{SAD}$\downarrow$ & \textbf{SA}$\uparrow$ \\
    \midrule
    \textbf{DemoFunGrasp} & 77.04 & 3.02 & 94.74  \\
    \hline
    \multicolumn{4}{l}{\textit{Component Ablation}} \\
    \quad - w/o afford reward & 78.28 & 4.60  & 95.02  \\
    \quad - w/o obj size clipping & \textbf{81.67} & 3.63 & 93.28\\
    \quad - w/o close reward & 76.44 & 3.79 & 90.58  \\
    \quad - w/o qpos reward & 74.38 & \textbf{2.98} & 95.36  \\
    \quad - w/o style disturbance & 58.67 & 3.61 & \textbf{100}\\
    \bottomrule
    \end{tabular}
\end{minipage}%
\hfill
\begin{minipage}{0.36\textwidth}
    \centering
    \includegraphics[width=0.95\linewidth, trim={0.cm, 0.cm, 0.cm, 0.cm}, clip]{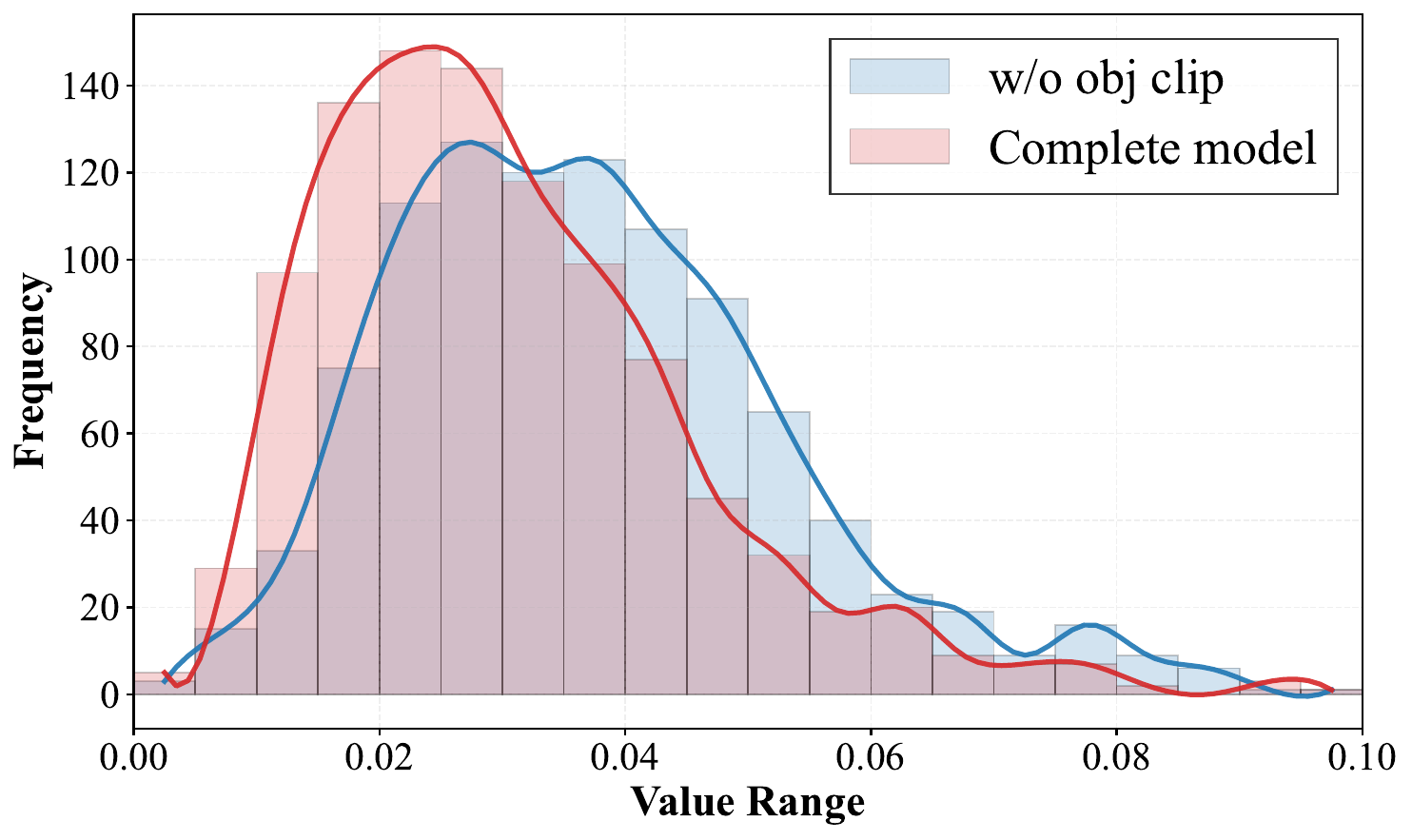}
    \vspace{-4mm}
    \captionof{figure}{\textbf{The affordance distance frequency of the complete model and the model without object size aware clipping (1000 samples).}}
    \label{fig:afford_dist_distribution}
\end{minipage}

With the additional filters—\emph{affordance distance$<$4\,cm} and \emph{style match}—the complete model achieves a balanced performance across grasp success, functional accuracy, and style fidelity, attaining the highest success rate of 60.55\%.

\subsection{Evaluation of Vision-based Policies}
\label{subsec:vision_exp_res}

We first evaluate our RGB-based policy in simulation on a set of objects (Appendix~\ref{subsec:Simulator_object_separation}) by randomly sampling affordances and styles to measure generalization capability. The grasp success rate and success affordance distance on the test set are 81.2\% and 3.79cm, respectively. Demonstrations are provided in Appendix~\ref{subsec:Demonstration_in_the_simulator}.

To validate zero-shot sim-to-real transfer, we deploy the RGB policy directly in the real world. Objects are categorized into three groups (Appendix~\ref{subsec:Real_world_object_separation}) based on their functionality and geometry. 
To generate functional affordance points automatically, we employ the state-of-the-art VLM \emph{Embodied-R1}~\cite{yuan2025embodiedr1}, with DemoFunGrasp used as the low-level control policy. The \textbf{Object Functional Grounding} prompt is used to extract affordance points reflecting object usage; details and comparisons to other VLMs are provided in Appendix~\ref{sec:VLM_Comparison_Results}.

We report the following metrics: 
\textbf{Intended Affordance Score (IAS)} = 1 if the grasped point matches the target affordance, 0 otherwise; 
\textbf{Intended Style Score (ISS)} = 1 if the final hand configuration matches the conditioned style, 0 otherwise. The results are shown in Table~\ref{tab:vision_exp_results}.

\begin{table}[!t]
    \caption{\textbf{Simulation and Real-world Results with Human-Annotated/VLM-Predicted Grasping Conditions.}}
   \label{tab:vision_exp_results}
    \vspace{-2mm}
    \centering
    \resizebox{0.9\textwidth}{!}{
        \begin{tabular}{c c c c>{\columncolor{yellow!30}}c>{\columncolor{blue!5}}c>{\columncolor{pink!30}}c>{\columncolor{yellow!30}}c>{\columncolor{blue!5}}c>{\columncolor{pink!30}}c}
        \toprule
        \multirow{3}{*}{\textbf{Metrics}} &
        \multicolumn{3}{c|}{\textbf{Simulator (human)}} &
        \multicolumn{3}{c|}{\textbf{Real World (human)}} &
        \multicolumn{3}{c}{\textbf{Real World (VLM)}} \\
        \cmidrule(r){2-4}\cmidrule(l){5-7}\cmidrule(l){8-10}
         & \textbf{Food} & \textbf{Kitchen} & \textbf{Tool}
         & \textbf{Daily item} & \textbf{Small tool} & \textbf{Large tool}
         & \textbf{Daily item} & \textbf{Small tool} & \textbf{Large tool} \\
        \midrule
        GSR$\uparrow$ & 22/25 & 18/25 & 20/25 & 11/15 & 11/15 & 10/15 & 12/15 & 10/15 & 7/15 \\
        IAS$\uparrow$ & 0.96 & 0.86 & 0.80 & 0.80 & 0.93 & 0.80 & 0.87 & 0.67 & 0.40 \\
        ISS$\uparrow$ & 1.00 & 0.82 & 0.92 & 0.80 & 1.00 & 0.93 & 0.93 & 0.87 & 0.87 \\
        \bottomrule
        \end{tabular}
    }
\end{table}

In simulation, kitchen objects yield lower performance than food items, primarily due to small or thin objects (e.g., forks) occluding the camera view when the hand approaches. 
In real-world experiments, DemoFunGrasp achieves 71\% GSR for human-chosen affordances and 64\% for VLM-predicted affordances, demonstrating robustness in functional grasp generation. Main failure modes include: 
(i) affordance regions too small for precise grasping; 
(ii) objects outside the training distribution; and 
(iii) VLM predictions misaligned due to challenging object poses or inaccessible regions.

\begin{figure*}[t!]
  \centering
  \includegraphics[width=0.90\linewidth, trim=5 0 5 0, clip]{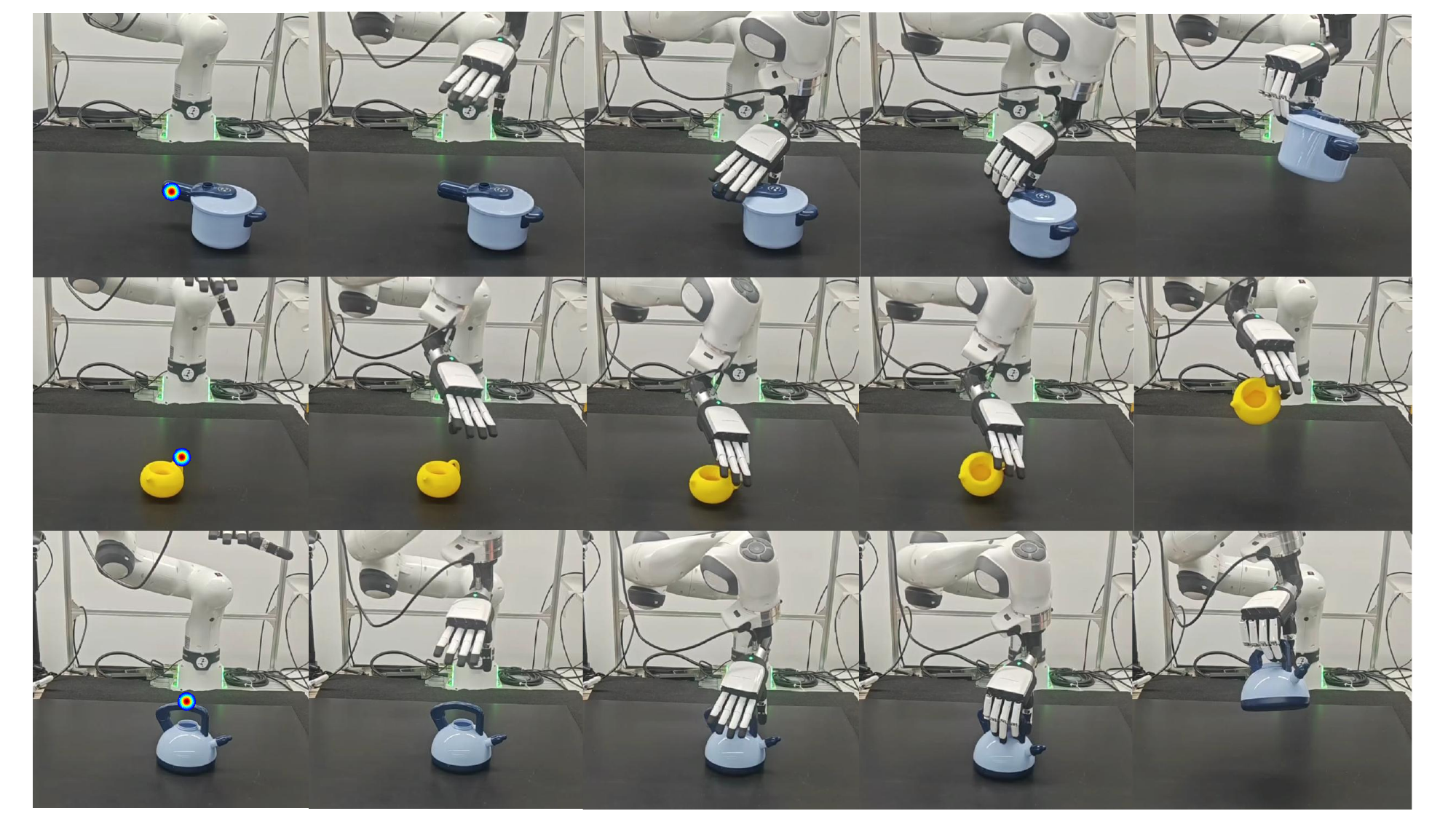}
  \caption{Real-world experimental demonstrations. The policy grasps a toy pot (first row), a teapot (second row), and a kettle (third row).}
  \label{fig:Real_world_experiments_demonstration}
  \vspace{-1em}
\end{figure*}

Overall, our method demonstrates strong performance on daily objects, achieving high success rates and accurate affordance and style adherence. Fig.~\ref{fig:Real_world_experiments_demonstration} illustrates representative trials, and additional demonstrations are provided in Appendix~\ref{subsec:Demonstration_in_the_real_world}.

\subsection{Additional Analysis} 

\textbf{Cross-Embodiment Evaluation.} We further assess the generality of our RL-based training framework across different robotic embodiments. Detailed experimental results and analysis are presented in Appendix~\ref{sec:Different_Embodiment_Experiment}. Our findings indicate that DemoFunGrasp consistently achieves high success rates when trained independently on robotic hands with varying degrees of freedom (DoF) and style capacities, demonstrating that the framework can be readily applied to diverse embodiments without any hyperparameter tuning.

\textbf{Base Algorithm Comparison.} In addition to reinforcement learning, we also evaluate several baseline strategies. The comparative results and corresponding analyses are provided in Appendix~\ref{sec:Other_Trials}.


\section{Conclusion and Limitations}

We present a framework for universal dexterous functional grasping by integrating style- and affordance-conditioned policies with an efficient RL method. Our approach achieves robust zero-shot sim-to-real performance and flexible adaptation to varied object geometries and functionalities. The integration of Vision-Language Models further enables autonomous and human-like planning and execution in real-world settings. We hope that DemoFunGrasp serves as a step toward more versatile and semantically aware dexterous manipulation systems.

However, the current performance of DemoFunGrasp still falls short of human capability, largely due to the absence of dynamic pre-grasp manipulation and in-hand adjustment. Although the policy can grasp diverse objects across different affordance regions and styles, its accuracy remains at the centimeter scale, indicating room for further improvement. Future work may explore learning dynamic adjustment policies and incorporating tactile sensing to improve accuracy.

\bibliography{ref}

\begin{thebibliography}{45}
\providecommand{\natexlab}[1]{#1}
\providecommand{\url}[1]{\texttt{#1}}
\expandafter\ifx\csname urlstyle\endcsname\relax
  \providecommand{\doi}[1]{doi: #1}\else
  \providecommand{\doi}{doi: \begingroup \urlstyle{rm}\Url}\fi

\bibitem[Agarwal et~al.(2023)Agarwal, Uppal, Shaw, and
  Pathak]{agarwal2023dexterous}
Ananye Agarwal, Shagun Uppal, Kenneth Shaw, and Deepak Pathak.
\newblock Dexterous functional grasping.
\newblock \emph{arXiv preprint arXiv:2312.02975}, 2023.

\bibitem[Ard{\'o}n et~al.(2019)Ard{\'o}n, Pairet, Petrick, Ramamoorthy, and
  Lohan]{ardon2019learning}
Paola Ard{\'o}n, Eric Pairet, Ronald~PA Petrick, Subramanian Ramamoorthy, and
  Katrin~S Lohan.
\newblock Learning grasp affordance reasoning through semantic relations.
\newblock \emph{IEEE Robotics and Automation Letters}, 4\penalty0 (4):\penalty0
  4571--4578, 2019.

\bibitem[Bahl et~al.(2023)Bahl, Mendonca, Chen, Jain, and
  Pathak]{bahl2023affordances}
Shikhar Bahl, Russell Mendonca, Lili Chen, Unnat Jain, and Deepak Pathak.
\newblock Affordances from human videos as a versatile representation for
  robotics.
\newblock In \emph{Proceedings of the IEEE/CVF Conference on Computer Vision
  and Pattern Recognition}, pp.\  13778--13790, 2023.

\bibitem[Bjorck et~al.(2025)Bjorck, Casta{\~n}eda, Cherniadev, Da, Ding, Fan,
  Fang, Fox, Hu, Huang, et~al.]{bjorck2025gr00t}
Johan Bjorck, Fernando Casta{\~n}eda, Nikita Cherniadev, Xingye Da, Runyu Ding,
  Linxi Fan, Yu~Fang, Dieter Fox, Fengyuan Hu, Spencer Huang, et~al.
\newblock Gr00t n1: An open foundation model for generalist humanoid robots.
\newblock \emph{arXiv preprint arXiv:2503.14734}, 2025.

\bibitem[Brahmbhatt et~al.(2019)Brahmbhatt, Ham, Kemp, and
  Hays]{brahmbhatt2019contactdb}
Samarth Brahmbhatt, Cusuh Ham, Charles~C Kemp, and James Hays.
\newblock Contactdb: Analyzing and predicting grasp contact via thermal
  imaging.
\newblock In \emph{Proceedings of the IEEE/CVF conference on computer vision
  and pattern recognition}, pp.\  8709--8719, 2019.

\bibitem[Calli et~al.(2015)Calli, Walsman, Singh, Srinivasa, Abbeel, and
  Dollar]{Calli_2015}
Berk Calli, Aaron Walsman, Arjun Singh, Siddhartha Srinivasa, Pieter Abbeel,
  and Aaron~M. Dollar.
\newblock Benchmarking in manipulation research: Using the yale-cmu-berkeley
  object and model set.
\newblock \emph{IEEE Robotics \& Automation Magazine}, 2015.

\bibitem[Chen et~al.(2025)Chen, Ke, Peng, and Wang]{chen2025dexonomy}
Jiayi Chen, Yubin Ke, Lin Peng, and He~Wang.
\newblock Dexonomy: Synthesizing all dexterous grasp types in a grasp taxonomy.
\newblock \emph{Robotics: Science and Systems}, 2025.

\bibitem[Detry et~al.(2011)Detry, Kraft, Kroemer, Bodenhagen, Peters,
  Kr{\"u}ger, and Piater]{detry2011learning}
Renaud Detry, Dirk Kraft, Oliver Kroemer, Leon Bodenhagen, Jan Peters, Norbert
  Kr{\"u}ger, and Justus Piater.
\newblock Learning grasp affordance densities.
\newblock \emph{Paladyn}, 2\penalty0 (1):\penalty0 1--17, 2011.

\bibitem[Duan et~al.(2021)Duan, Wang, Huang, Xu, Wei, and
  Shen]{duan2021robotics}
Haonan Duan, Peng Wang, Yayu Huang, Guangyun Xu, Wei Wei, and Xiaofei Shen.
\newblock Robotics dexterous grasping: The methods based on point cloud and
  deep learning.
\newblock \emph{Frontiers in Neurorobotics}, 15:\penalty0 658280, 2021.

\bibitem[Feix et~al.(2015)Feix, Romero, Schmiedmayer, Dollar, and
  Kragic]{feix2015grasp}
Thomas Feix, Javier Romero, Heinz-Bodo Schmiedmayer, Aaron~M Dollar, and Danica
  Kragic.
\newblock The grasp taxonomy of human grasp types.
\newblock \emph{IEEE Transactions on human-machine systems}, 46\penalty0
  (1):\penalty0 66--77, 2015.

\bibitem[Grady et~al.(2021)Grady, Tang, Twigg, Vo, Brahmbhatt, and
  Kemp]{grady2021contactopt}
Patrick Grady, Chengcheng Tang, Christopher~D. Twigg, Minh Vo, Samarth
  Brahmbhatt, and Charles~C. Kemp.
\newblock {ContactOpt}: Optimizing contact to improve grasps.
\newblock In \emph{Conference on Computer Vision and Pattern Recognition
  (CVPR)}, 2021.

\bibitem[He et~al.(2025)He, Li, Yu, Qi, Zhang, Chen, Zhang, Zhang, Yi, and
  Wang]{dexvlg25}
Jiawei He, Danshi Li, Xinqiang Yu, Zekun Qi, Wenyao Zhang, Jiayi Chen,
  Zhaoxiang Zhang, Zhizheng Zhang, Li~Yi, and He~Wang.
\newblock Dexvlg: Dexterous vision-language-grasp model at scale.
\newblock \emph{arXiv preprint arXiv:2507.02747}, 2025.

\bibitem[Hu et~al.(2025)Hu, Huang, Lee, Yang, Zheng, and Li]{hu2025dexterous}
Wenbin Hu, Bidan Huang, Wang~Wei Lee, Sicheng Yang, Yu~Zheng, and Zhibin Li.
\newblock Dexterous in-hand manipulation of slender cylindrical objects through
  deep reinforcement learning with tactile sensing.
\newblock \emph{Robotics and Autonomous Systems}, 186:\penalty0 104904, 2025.

\bibitem[Huang et~al.(2025)Huang, Zhang, Wu, Christen, and
  Song]{huang2025fungrasp}
Linyi Huang, Hui Zhang, Zijian Wu, Sammy Christen, and Jie Song.
\newblock Fungrasp: functional grasping for diverse dexterous hands.
\newblock \emph{IEEE Robotics and Automation Letters}, 2025.

\bibitem[Huang et~al.(2024)Huang, Yuan, Fu, and Lu]{huang2024efficient}
Ziye Huang, Haoqi Yuan, Yuhui Fu, and Zongqing Lu.
\newblock Efficient residual learning with mixture-of-experts for universal
  dexterous grasping.
\newblock \emph{arXiv preprint arXiv:2410.02475}, 2024.

\bibitem[Jian et~al.(2023)Jian, Liu, Li, Hu, and Liu]{Jian_2023_ICCV}
Juntao Jian, Xiuping Liu, Manyi Li, Ruizhen Hu, and Jian Liu.
\newblock Affordpose: A large-scale dataset of hand-object interactions with
  affordance-driven hand pose.
\newblock In \emph{Proceedings of the IEEE/CVF International Conference on
  Computer Vision (ICCV)}, pp.\  14713--14724, October 2023.

\bibitem[Kokic et~al.(2017)Kokic, Stork, Haustein, and
  Kragic]{kokic2017affordance}
Mia Kokic, Johannes~A Stork, Joshua~A Haustein, and Danica Kragic.
\newblock Affordance detection for task-specific grasping using deep learning.
\newblock In \emph{2017 IEEE-RAS 17th International Conference on Humanoid
  Robotics (Humanoids)}, pp.\  91--98. IEEE, 2017.

\bibitem[Lin et~al.(2025)Lin, Sachdev, Fan, Malik, and Zhu]{lin2025sim}
Toru Lin, Kartik Sachdev, Linxi Fan, Jitendra Malik, and Yuke Zhu.
\newblock Sim-to-real reinforcement learning for vision-based dexterous
  manipulation on humanoids.
\newblock \emph{arXiv preprint arXiv:2502.20396}, 2025.

\bibitem[Liu et~al.(2022)Liu, Liu, Jiang, Lyu, Wan, Shen, Liang, Fu, Wang, and
  Yi]{liu2022hoi4d}
Yunze Liu, Yun Liu, Che Jiang, Kangbo Lyu, Weikang Wan, Hao Shen, Boqiang
  Liang, Zhoujie Fu, He~Wang, and Li~Yi.
\newblock Hoi4d: A 4d egocentric dataset for category-level human-object
  interaction.
\newblock In \emph{Proceedings of the IEEE/CVF Conference on Computer Vision
  and Pattern Recognition}, pp.\  21013--21022, 2022.

\bibitem[Luo et~al.(2025)Luo, Xu, Wu, and Levine]{luo2025precise}
Jianlan Luo, Charles Xu, Jeffrey Wu, and Sergey Levine.
\newblock Precise and dexterous robotic manipulation via human-in-the-loop
  reinforcement learning.
\newblock \emph{Science Robotics}, 10\penalty0 (105):\penalty0 eads5033, 2025.

\bibitem[Makoviychuk et~al.(2021)Makoviychuk, Wawrzyniak, Guo, Lu, Storey,
  Macklin, Hoeller, Rudin, Allshire, Handa, et~al.]{isaacgym}
Viktor Makoviychuk, Lukasz Wawrzyniak, Yunrong Guo, Michelle Lu, Kier Storey,
  Miles Macklin, David Hoeller, Nikita Rudin, Arthur Allshire, Ankur Handa,
  et~al.
\newblock Isaac gym: High performance gpu-based physics simulation for robot
  learning.
\newblock \emph{arXiv preprint arXiv:2108.10470}, 2021.

\bibitem[Mandikal \& Grauman(2021)Mandikal and Grauman]{mandikal2021learning}
Priyanka Mandikal and Kristen Grauman.
\newblock Learning dexterous grasping with object-centric visual affordances.
\newblock In \emph{2021 IEEE international conference on robotics and
  automation (ICRA)}, pp.\  6169--6176. IEEE, 2021.

\bibitem[Mandikal \& Grauman(2022)Mandikal and Grauman]{mandikal2022dexvip}
Priyanka Mandikal and Kristen Grauman.
\newblock Dexvip: Learning dexterous grasping with human hand pose priors from
  video.
\newblock In \emph{Conference on Robot Learning}, pp.\  651--661. PMLR, 2022.

\bibitem[OpenAI et~al.(2018)OpenAI, Andrychowicz, Baker, Chociej, Józefowicz,
  McGrew, Pachocki, Petron, Plappert, Powell, Ray, Schneider, Sidor, Tobin,
  Welinder, Weng, and Zaremba]{openai2018learning}
OpenAI, Marcin Andrychowicz, Bowen Baker, Maciek Chociej, Rafał Józefowicz,
  Bob McGrew, Jakub Pachocki, Arthur Petron, Matthias Plappert, Glenn Powell,
  Alex Ray, Jonas Schneider, Szymon Sidor, Josh Tobin, Peter Welinder, Lilian
  Weng, and Wojciech Zaremba.
\newblock Learning dexterous in-hand manipulation.
\newblock \emph{CoRR}, 2018.
\newblock URL \url{http://arxiv.org/abs/1808.00177}.

\bibitem[Popov et~al.(2017)Popov, Heess, Lillicrap, Hafner, Barth-Maron,
  Vecerik, Lampe, Tassa, Erez, and Riedmiller]{popov2017data}
Ivaylo Popov, Nicolas Heess, Timothy Lillicrap, Roland Hafner, Gabriel
  Barth-Maron, Matej Vecerik, Thomas Lampe, Yuval Tassa, Tom Erez, and Martin
  Riedmiller.
\newblock Data-efficient deep reinforcement learning for dexterous
  manipulation.
\newblock \emph{arXiv preprint arXiv:1704.03073}, 2017.

\bibitem[Rajeswaran et~al.(2017)Rajeswaran, Kumar, Gupta, Vezzani, Schulman,
  Todorov, and Levine]{rajeswaran2017learning}
Aravind Rajeswaran, Vikash Kumar, Abhishek Gupta, Giulia Vezzani, John
  Schulman, Emanuel Todorov, and Sergey Levine.
\newblock Learning complex dexterous manipulation with deep reinforcement
  learning and demonstrations.
\newblock \emph{arXiv preprint arXiv:1709.10087}, 2017.

\bibitem[Schulman et~al.(2017)Schulman, Wolski, Dhariwal, Radford, and
  Klimov]{ppo}
John Schulman, Filip Wolski, Prafulla Dhariwal, Alec Radford, and Oleg Klimov.
\newblock Proximal policy optimization algorithms.
\newblock \emph{arXiv preprint arXiv:1707.06347}, 2017.

\bibitem[Song et~al.(2015)Song, Fritz, Goehring, and Darrell]{song2015learning}
Hyun~Oh Song, Mario Fritz, Daniel Goehring, and Trevor Darrell.
\newblock Learning to detect visual grasp affordance.
\newblock \emph{IEEE Transactions on Automation Science and Engineering},
  13\penalty0 (2):\penalty0 798--809, 2015.

\bibitem[Sun et~al.(2022)Sun, Amatova, and Chen]{9812388}
Yu~Sun, Eliza Amatova, and Tianze Chen.
\newblock Multi-object grasping - types and taxonomy.
\newblock In \emph{2022 International Conference on Robotics and Automation
  (ICRA)}, pp.\  777--783, 2022.
\newblock \doi{10.1109/ICRA46639.2022.9812388}.

\bibitem[Wan et~al.(2023)Wan, Geng, Liu, Shan, Yang, Yi, and
  Wang]{wan2023unidexgrasp++}
Weikang Wan, Haoran Geng, Yun Liu, Zikang Shan, Yaodong Yang, Li~Yi, and
  He~Wang.
\newblock Unidexgrasp++: Improving dexterous grasping policy learning via
  geometry-aware curriculum and iterative generalist-specialist learning.
\newblock In \emph{Proceedings of the IEEE/CVF International Conference on
  Computer Vision}, pp.\  3891--3902, 2023.

\bibitem[Wang et~al.(2022)Wang, Zhang, Chen, Xu, Li, Liu, and
  Wang]{wang2022dexgraspnet}
Ruicheng Wang, Jialiang Zhang, Jiayi Chen, Yinzhen Xu, Puhao Li, Tengyu Liu,
  and He~Wang.
\newblock Dexgraspnet: A large-scale robotic dexterous grasp dataset for
  general objects based on simulation.
\newblock \emph{arXiv preprint arXiv:2210.02697}, 2022.

\bibitem[Wei et~al.(2024)Wei, Jiang, Xing, Tan, Wu, Li, Cutkosky, and
  Zheng]{wei2024grasp}
Yi-Lin Wei, Jian-Jian Jiang, Chengyi Xing, Xian-Tuo Tan, Xiao-Ming Wu, Hao Li,
  Mark Cutkosky, and Wei-Shi Zheng.
\newblock Grasp as you say: Language-guided dexterous grasp generation.
\newblock \emph{Advances in Neural Information Processing Systems},
  37:\penalty0 46881--46907, 2024.

\bibitem[Wei et~al.(2025{\natexlab{a}})Wei, Lin, Lin, Jiang, Wu, Zeng, and
  Zheng]{wei2025afforddexgrasp}
Yi-Lin Wei, Mu~Lin, Yuhao Lin, Jian-Jian Jiang, Xiao-Ming Wu, Ling-An Zeng, and
  Wei-Shi Zheng.
\newblock Afforddexgrasp: Open-set language-guided dexterous grasp with
  generalizable-instructive affordance.
\newblock \emph{arXiv preprint arXiv:2503.07360}, 2025{\natexlab{a}}.

\bibitem[Wei et~al.(2025{\natexlab{b}})Wei, Luo, Lin, Lin, Liang, Chen, and
  Zheng]{wei2025omnidexgrasp}
Yi-Lin Wei, Zhexi Luo, Yuhao Lin, Mu~Lin, Zhizhao Liang, Shuoyu Chen, and
  Wei-Shi Zheng.
\newblock Omnidexgrasp: Generalizable dexterous grasping via foundation model
  and force feedback.
\newblock \emph{arXiv preprint arXiv:2510.23119}, 2025{\natexlab{b}}.

\bibitem[Wu et~al.(2024{\natexlab{a}})Wu, Zhu, Lin, and Sun]{wu2024cross}
Rina Wu, Tianqiang Zhu, Xiangbo Lin, and Yi~Sun.
\newblock Cross-category functional grasp transfer.
\newblock \emph{IEEE Robotics and Automation Letters}, 2024{\natexlab{a}}.

\bibitem[Wu et~al.(2024{\natexlab{b}})Wu, Gan, Wu, Cheng, Yang, Zhu, and
  Dong]{wu2024dexterous}
Tianhao Wu, Yunchong Gan, Mingdong Wu, Jingbo Cheng, Yaodong Yang, Yixin Zhu,
  and Hao Dong.
\newblock Dexterous functional pre-grasp manipulation with diffusion policy.
\newblock \emph{arXiv preprint arXiv:2403.12421}, 2024{\natexlab{b}}.

\bibitem[Xu et~al.(2023)Xu, Wan, Zhang, Liu, Shan, Shen, Wang, Geng, Weng,
  Chen, et~al.]{xu2023unidexgrasp}
Yinzhen Xu, Weikang Wan, Jialiang Zhang, Haoran Liu, Zikang Shan, Hao Shen,
  Ruicheng Wang, Haoran Geng, Yijia Weng, Jiayi Chen, et~al.
\newblock Unidexgrasp: Universal robotic dexterous grasping via learning
  diverse proposal generation and goal-conditioned policy.
\newblock In \emph{Proceedings of the IEEE/CVF Conference on Computer Vision
  and Pattern Recognition}, pp.\  4737--4746, 2023.

\bibitem[Yuan et~al.(2024)Yuan, Zhou, Fu, and Lu]{yuan2024cross}
Haoqi Yuan, Bohan Zhou, Yuhui Fu, and Zongqing Lu.
\newblock Cross-embodiment dexterous grasping with reinforcement learning.
\newblock \emph{arXiv preprint arXiv:2410.02479}, 2024.

\bibitem[Yuan et~al.(2025{\natexlab{a}})Yuan, Huang, Wang, Mao, Xu, and
  Lu]{yuan2025demograsp}
Haoqi Yuan, Ziye Huang, Ye~Wang, Chuan Mao, Chaoyi Xu, and Zongqing Lu.
\newblock Demograsp: Universal dexterous grasping from a single demonstration.
\newblock \emph{arXiv preprint arXiv:2509.22149}, 2025{\natexlab{a}}.

\bibitem[Yuan et~al.(2025{\natexlab{b}})Yuan, Cui, Huang, Chen, Ni, Dong, Li,
  Zheng, and Hao]{yuan2025embodiedr1}
Yifu Yuan, Haiqin Cui, Yaoting Huang, Yibin Chen, Fei Ni, Zibin Dong, Pengyi
  Li, Yan Zheng, and Jianye Hao.
\newblock Embodied-r1: Reinforced embodied reasoning for general robotic
  manipulation.
\newblock \emph{arXiv preprint arXiv:2508.13998}, 2025{\natexlab{b}}.

\bibitem[Zhang et~al.(2025)Zhang, Wu, Huang, Christen, and
  Song]{zhang2025robustdexgrasp}
Hui Zhang, Zijian Wu, Linyi Huang, Sammy Christen, and Jie Song.
\newblock Robustdexgrasp: Robust dexterous grasping of general objects.
\newblock \emph{arXiv preprint arXiv:2504.05287}, 2025.

\bibitem[Zhao et~al.(2023)Zhao, Kumar, Levine, and Finn]{zhao2023learning}
Tony~Z Zhao, Vikash Kumar, Sergey Levine, and Chelsea Finn.
\newblock Learning fine-grained bimanual manipulation with low-cost hardware.
\newblock \emph{arXiv preprint arXiv:2304.13705}, 2023.

\bibitem[Zhong et~al.(2025)Zhong, Huang, Li, Zhang, Chen, Guan, Zeng, Lui, Ye,
  Liang, et~al.]{zhong2025dexgraspvla}
Yifan Zhong, Xuchuan Huang, Ruochong Li, Ceyao Zhang, Zhang Chen, Tianrui Guan,
  Fanlian Zeng, Ka~Num Lui, Yuyao Ye, Yitao Liang, et~al.
\newblock Dexgraspvla: A vision-language-action framework towards general
  dexterous grasping.
\newblock \emph{arXiv preprint arXiv:2502.20900}, 2025.

\bibitem[Zhou et~al.(2024)Zhou, Yuan, Fu, and Lu]{zhou2024learning}
Bohan Zhou, Haoqi Yuan, Yuhui Fu, and Zongqing Lu.
\newblock Learning diverse bimanual dexterous manipulation skills from human
  demonstrations.
\newblock \emph{arXiv preprint arXiv:2410.02477}, 2024.

\bibitem[Zhu et~al.(2025)Zhu, Bai, Xiang, Cai, Chen, Li, Wang, Dong, Yang, Fan,
  et~al.]{zhu2025dexflywheel}
Kefei Zhu, Fengshuo Bai, YuanHao Xiang, Yishuai Cai, Xinglin Chen, Ruochong Li,
  Xingtao Wang, Hao Dong, Yaodong Yang, Xiaopeng Fan, et~al.
\newblock Dexflywheel: A scalable and self-improving data generation framework
  for dexterous manipulation.
\newblock \emph{arXiv preprint arXiv:2509.23829}, 2025.

\end{thebibliography}
\bibliographystyle{iclr2026_conference}

\newpage
\appendix

\section{Hardware Setup}
\label{sec:The_Robot_workspace}

\begin{figure}[ht!]
\centering
    \begin{minipage}{0.5\textwidth}
    \small

    The real-world hardware setup is shown in Fig.~\ref{fig:exp_setting}. 
We use a Franka 3 arm paired with an Inspire robotic right hand. 
Objects are randomly placed within a \(0.3 \times 0.3\,\mathrm{m}\) workspace region. 
To obtain visual observations, we deploy two Intel RealSense D435i cameras positioned to provide complementary viewpoints of the workspace. 
We perform standard hand--eye calibration to accurately estimate each cameras' extrinsic parameters, and these calibrated parameters are imported into the simulation environment to ensure consistency between real and simulated camera poses. 
The RGB images captured by the cameras are resized to \(256 \times 256\) and used as inputs to our vision-based policy. 
This resolution is selected because it offers a favorable trade-off between detail preservation and system efficiency: compared with higher-resolution inputs, \(256 \times 256\) images significantly reduce data-transfer latency and computational load while retaining the necessary geometric and semantic cues for reliable perception and control.
    \end{minipage}%
\hfill
    \begin{minipage}{0.42\textwidth}
    \centering
    \includegraphics[width=\textwidth]{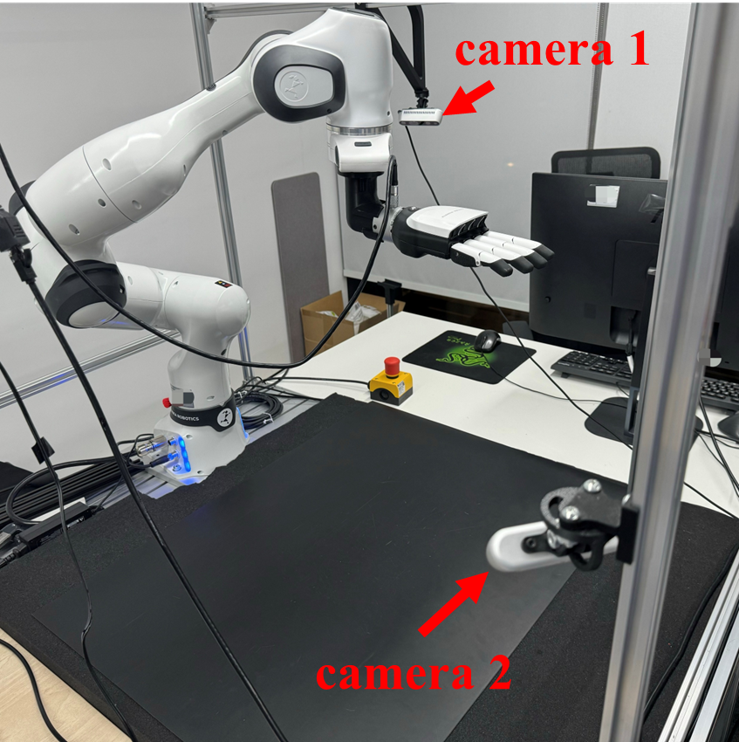}
    \caption{The real-world hardware setup.}
    \label{fig:exp_setting}
    \end{minipage}
\end{figure}

\section{Objects Used in Experiments}
\label{sec:Objects_in_the_experiment}

\subsection{State-based Training Dataset}

In the \textbf{state-based configuration}, we utilize three datasets to comprehensively evaluate our method across multiple dimensions:
\begin{itemize}
    \item \textbf{DexGraspNet~\cite{wang2022dexgraspnet}}: a large-scale dexterous grasping dataset containing over 3,200 diverse objects, serving as the primary benchmark for evaluating general grasp performance.
    \item \textbf{YCB~\cite{Calli_2015}}: a standard benchmark consisting of 75 everyday objects and tools, used to assess the ability of DemoFunGrasp to perform functionally diverse grasps.
    \item \textbf{AffordObj}: a dataset constructed for functional grasping tasks that require attending to specific object regions, derived from the YCB object set.
\end{itemize}

\begin{figure}[t!]
    \centering
    \begin{minipage}{0.45\textwidth}
        \centering
        \includegraphics[width=\linewidth]{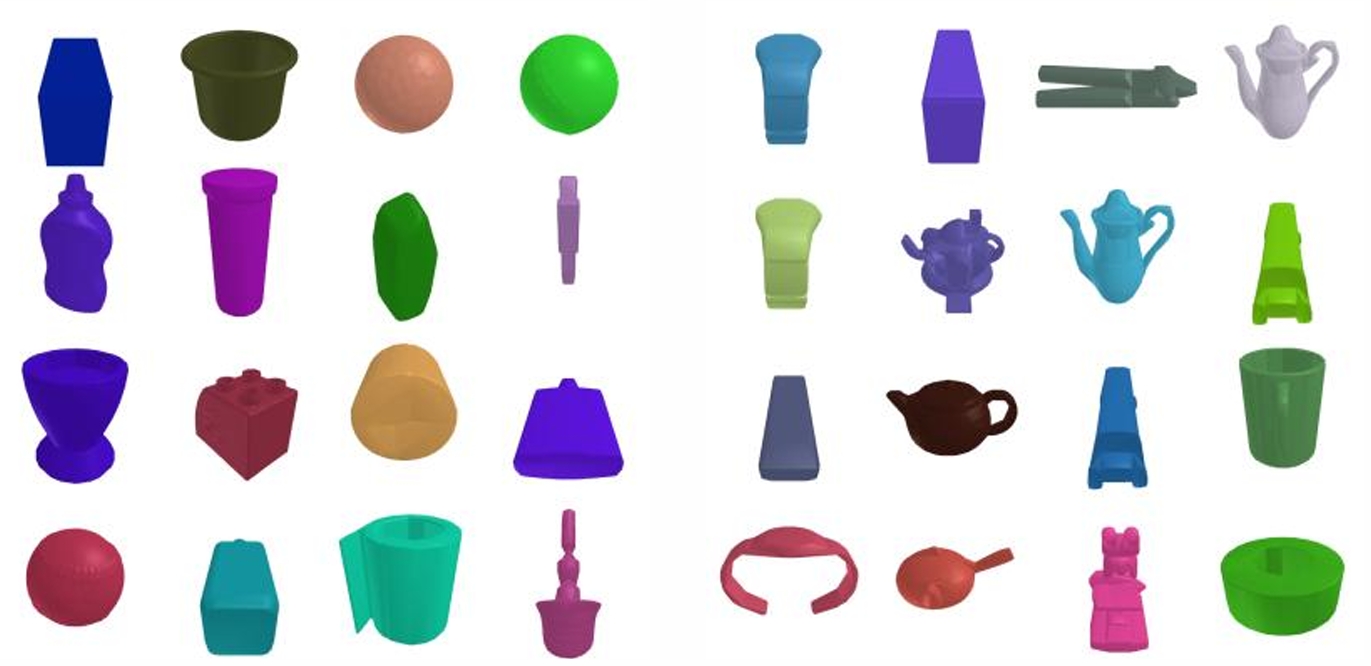}
        \caption{Objects used to train state-based policy in the simulator.}
        \label{fig:sim_train_obj}
    \end{minipage}
    \hfill
    \begin{minipage}{0.5\textwidth}
        \centering
        \includegraphics[width=\linewidth,keepaspectratio, trim=200 200 230 100, clip]{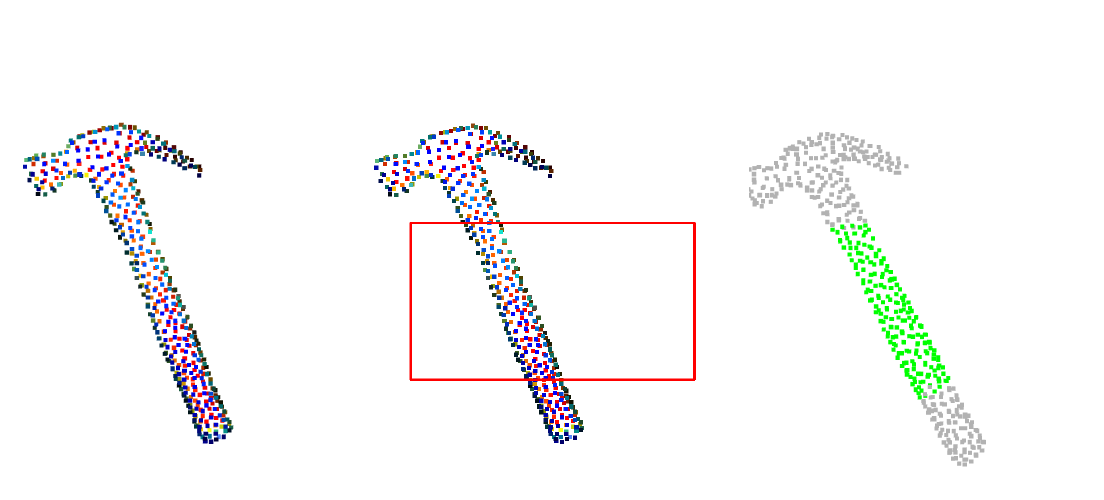}
        \caption{Pipeline for annotating affordance points on point clouds.}
        \label{fig:pcl_annotate}
    \end{minipage}
\end{figure}

For our training and evaluation split, we curate a mixed dataset of 175 objects sourced from DexGraspNet and YCB. This combined dataset provides a broader and more varied object distribution than either dataset alone. A small subset of the mixed dataset is shown in Fig.~\ref{fig:sim_train_obj}.

\subsection{Dataset Processing}

To enable more accurate sampling of affordance points on objects, we adopt a manual annotation pipeline (Fig.~\ref{fig:pcl_annotate}). We additionally leverage the AffordPose~\cite{Jian_2023_ICCV} dataset to sample affordance points for training the state-based model. Our experiments indicate that human-preferred functional regions do not necessarily correspond to regions that are easy for the robot to grasp. Consequently, conditioning solely on human-labeled functional regions does not improve either the training success rate or the affordance accuracy of the state-based model.

Our main framework requires providing the model with high-level semantic information. Experiments show that, after training on the annotated dataset and distilling the model into a vision-based policy \emph{without} explicit affordance conditioning, the student policy can still generalize to seen category objects by grasping their functional regions autonomously—without human or VLM guidance. This demonstrates that our method is capable of acquiring semantic understanding when scaled to a sufficiently large annotated dataset.


\subsection{Simulation Object Categorization}
\label{subsec:Simulator_object_separation}

\begin{figure}[t!]
    \centering
    \begin{minipage}{0.45\textwidth}
        \centering
        \includegraphics[width=0.8\textwidth]{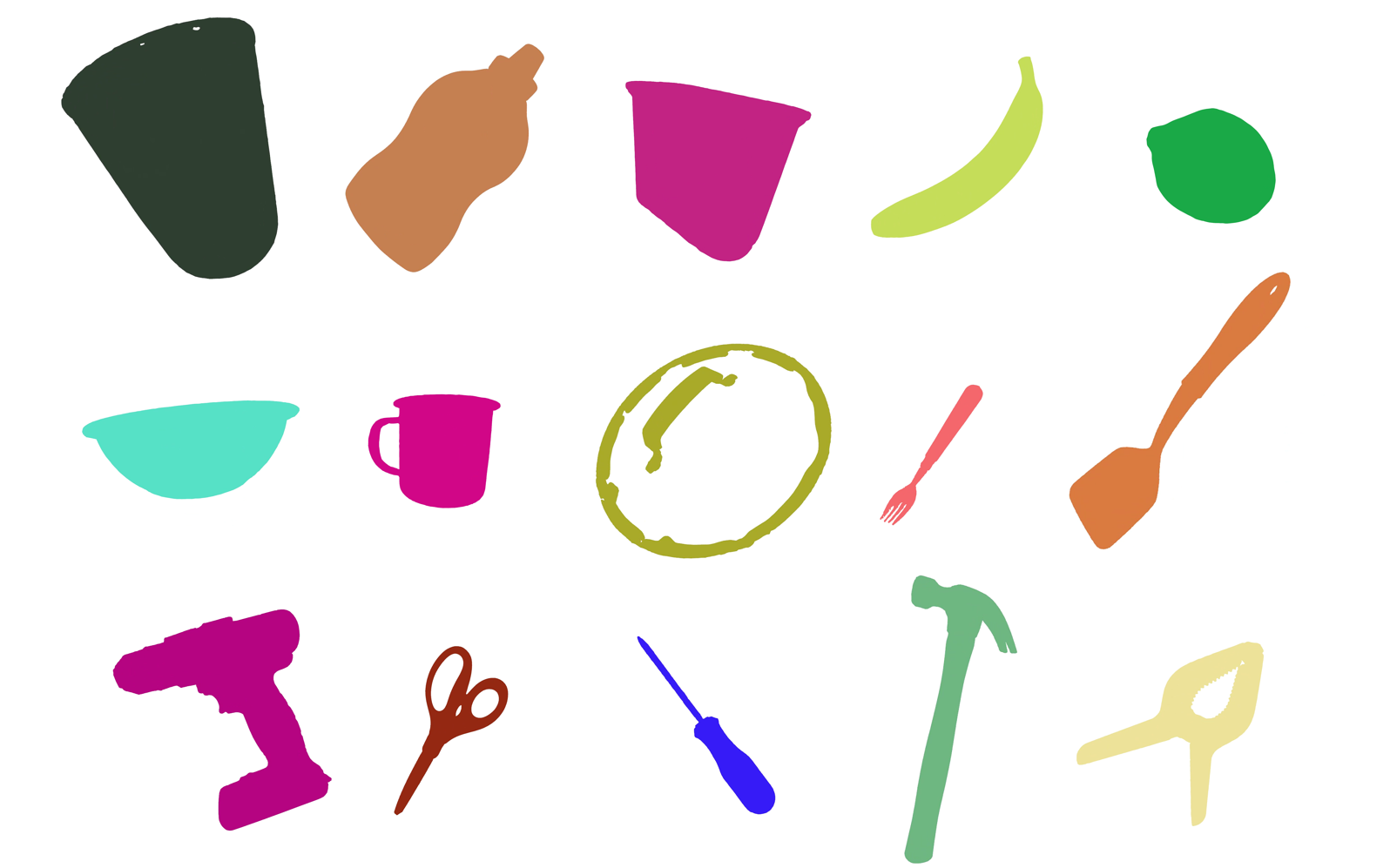}
          \caption{Objects used to evaluate the vision-based policy in simulation.}
          \label{fig:sim_obj_sep}
    \end{minipage}
    \hfill
    \begin{minipage}{0.5\textwidth}
        \centering
        \includegraphics[width=0.95\textwidth,
          keepaspectratio, trim=20 70 45 10, clip]{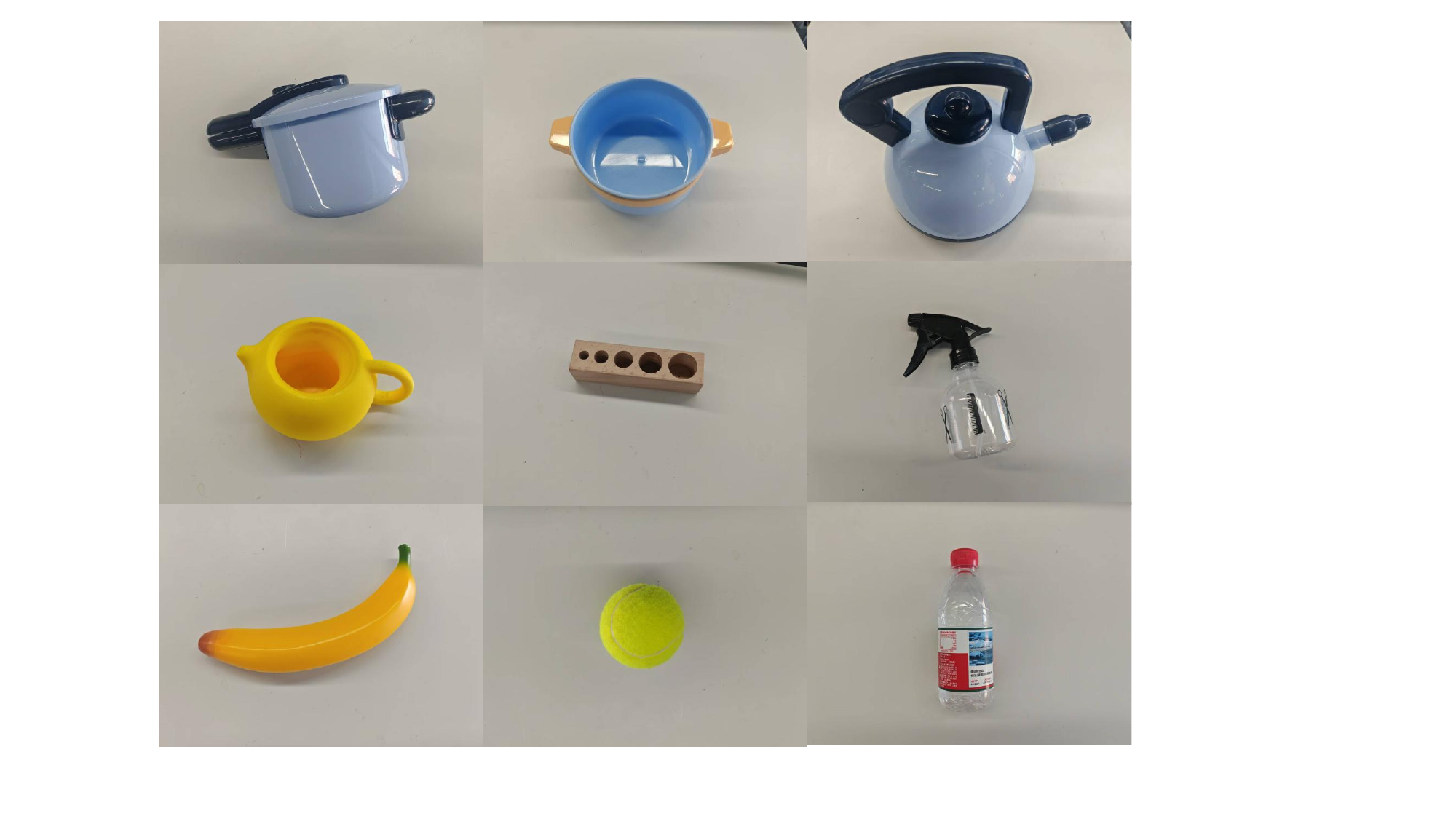}
          \caption{Objects used to evaluate the vision-based policy in real-world experiments.}
          \label{fig:real_obj_sep}
    \end{minipage}
\end{figure}

In the simulator, after training on a large and diverse set of objects, we evaluate the vision-based policy under human guidance by specifying the desired affordance and grasping style. To systematically assess generalization, we categorize objects into three shape-based classes: \textit{food items}, \textit{kitchen items}, and \textit{tools}. For each category, we select five representative objects for evaluation (Fig.~\ref{fig:sim_obj_sep}).

\subsection{Real-World Object Categorization}
\label{subsec:Real_world_object_separation}

Objects used in the real-world test set are shown in Fig.~\ref{fig:real_obj_sep} and categorized as follows:
\begin{itemize}
    \item \textbf{Daily Items}: everyday objects with no clearly defined affordance region, such as balls and bananas.
    \item \textbf{Small Tools}: compact tools including spray bottles and small teapots.
    \item \textbf{Large Tools}: larger household tools such as bowls and kettles.
\end{itemize}
\section{Cross-Embodiment Evaluation}
\label{sec:Different_Embodiment_Experiment}

We evaluate DemoFunGrasp across three robotic hand embodiments.  
For the Shadow Hand, grasping style priors are derived from the Dexonomy~\cite{chen2025dexonomy} dataset,  
whereas for the Inspire Hand and Wuji Hand, styles are manually defined via joint tuning.  
Both sources of style initialization require minimal manual effort while effectively achieving high Grasp Success Rate (GSR) and low Success Affordance Distance (SAD).  

Comprehensive results for the state-based evaluation are presented in Table~\ref{tab:cross_embodiment}. The Inspire Hand achieves the highest success rate and the lowest affordance distance, likely due to its lower degrees of freedom (DoF), which simplify optimization.

\begin{table}[ht!]
\centering
\caption{\textbf{Cross-Embodiment Evaluation of DemoFunGrasp.}}
\label{tab:cross_embodiment}
\begin{tabular}{lcc|cc}
\toprule
\textbf{Hand} & \textbf{DoF} & \textbf{Styles} & \textbf{GSR}$\uparrow$  & \textbf{SAD}$\downarrow$ \\
\midrule
Inspire Hand & 6  & 4 & \textbf{87.85} & \textbf{2.66} \\
Shadow Hand  & 22 & 9 & 77.04 & 3.02  \\
Wuji Hand    & 20 & 9 & 77.09 & 2.74 \\
\bottomrule
\end{tabular}
\end{table}


\section{The VLM Planner}
\label{sec:VLM_Comparison_Results}

Prompt for \textbf{ChatGPT} and \textbf{Gemini 2.5 pro}: 
\begin{tcolorbox}
``template'': (
``Please provide the 2D point coordinate of the region this sentence describes: \{instruction\}.'' ``The input image size is 256×256 pixels.''
``Generate 4 candidate points and select the best one for grasp affordance.'' ``The results are presented in a format<point>[x,y]</point>.'' ``You FIRST think about the reasoning process as an internal monologue and then provide the final answer.''
``The reasoning process and answer are enclosed within <think></think> and <answer></answer> tags.'' ``The answer consists of only one coordinate point, with the overall format being: <think> reasoning process here </think><answer><point>[x,y]</point></answer>.'' ``Important: the point must lie on the object, not on the background or table surface.''
),

``description'': ``Object Affordance Grounding - Locating the 2D coordinates of specified object regions based on descriptions.''
\end{tcolorbox}

Prompt for \textbf{Embodied-R1}: 
\begin{tcolorbox}
``template'': (``Please provide the 2D points coordinate of the region this sentence describes: \{instruction\}.'' ``The results are presented in a format <point>[[x1,y1], [x2,y2], ...]</point>.'' ``You FIRST think about the reasoning process as an internal monologue and then provide the final answer.'' ``The reasoning process and answer are enclosed within<think></think>and<answer></answer> tags'' ``The answer consists only of several coordinate points, with the overall format being: <think> reasoning process here </think><answer><point>[[x1,y1], [x2,y2],...]</point></answer>''
),

"description": "Object Affordance Grounding - Locating the 2D coordinates of specified object regions based on descriptions."
\end{tcolorbox}

Instruction:
\begin{tcolorbox}
Grasp the object on the table by identifying the optimal affordance region, and return the coordinates of the reasoning points.
\end{tcolorbox}

\begin{figure}[ht!] 
    \centering 
    \includegraphics[width=0.95\textwidth]{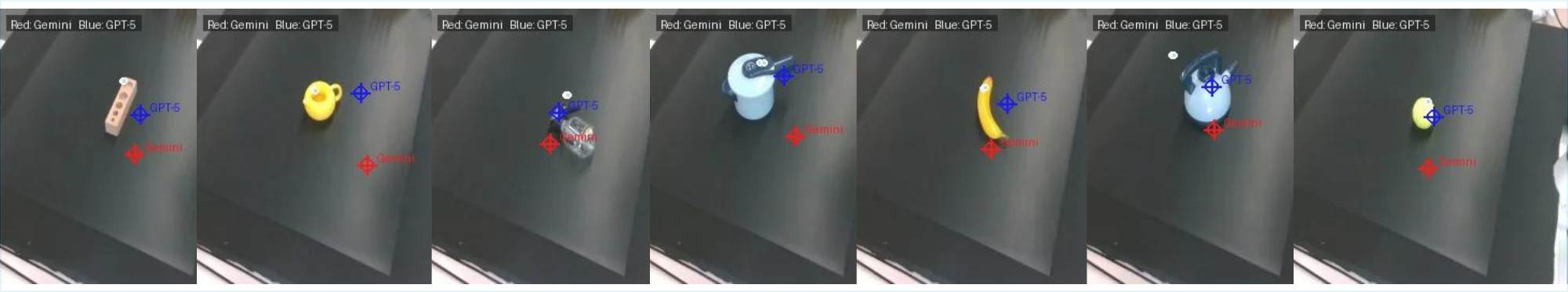} 
    \caption{Comparative evaluation of Embodied-R1 (white points), Gemini 2.5 Pro (red points), and GPT-5 (blue points).} 
    \label{fig:vlm_pointing_comp} 
\end{figure}

The VLM-generated outputs are shown in Fig.~\ref{fig:vlm_pointing_comp}. While Gemini 2.5 Pro and GPT-5 demonstrate strong reasoning capabilities and can produce logically coherent interpretations of tabletop scenes, they consistently fail to generate precise point coordinates. We hypothesize that this limitation stems from their insufficient modeling of pixel-level spatial information.

Our insight is that, to achieve a universal robotic manipulation policy, it is necessary to train a high-level “cognitive” model capable of long-horizon reasoning and task planning, while the low-level policy focuses primarily on ensuring execution robustness and stability.


\section{Additional Qualitative Results}

\subsection{Results in Simulation}
\label{subsec:Demonstration_in_the_simulator}

\begin{figure*}[ht!]
    \centering
    \includegraphics[width=1.0\textwidth, keepaspectratio, trim=10 50 10 20, clip]{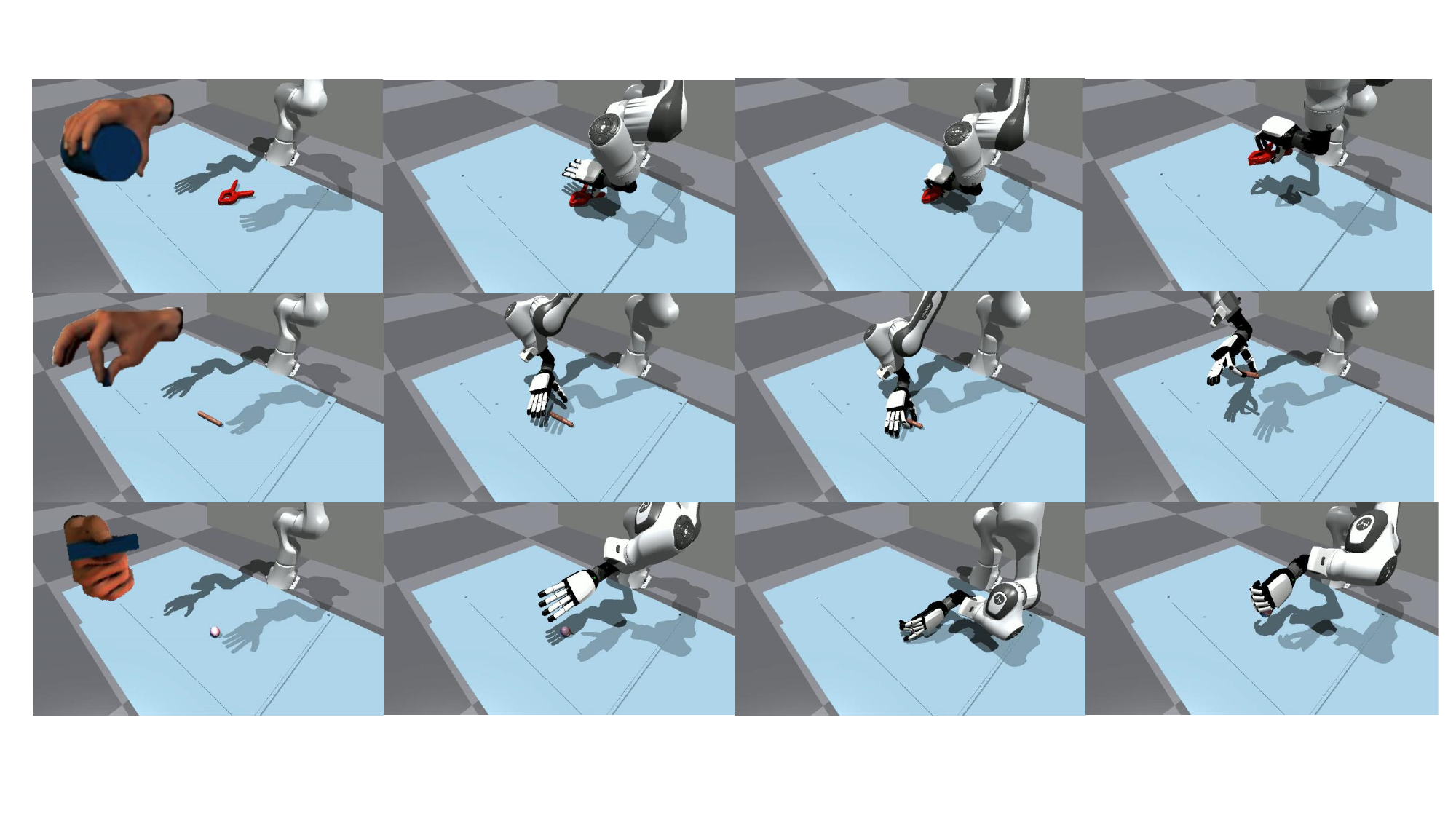}
    \vspace{-2em}
    \caption{Simulator video recordings of the vision-based policy across diverse objects and grasping styles.}
    \label{fig:sim_demo}
\end{figure*}

Fig.~\ref{fig:sim_demo} presents recordings of the vision-based policy in simulation, demonstrating its versatility across a range of object types and grasping scenarios. In addition to executing grasps with a specified hand style, our method effectively handles small, thin, or fragile objects, as well as objects prone to rolling or instability. These demonstrations highlight the policy's ability to adapt to challenging object geometries and physical dynamics, showcasing its generalization capability within the simulator.

\subsection{Real-World Results}
\label{subsec:Demonstration_in_the_real_world}

\begin{figure*}[h!]
    \centering
    \includegraphics[width=1.0\textwidth, keepaspectratio, trim=220 0 220 0, clip]{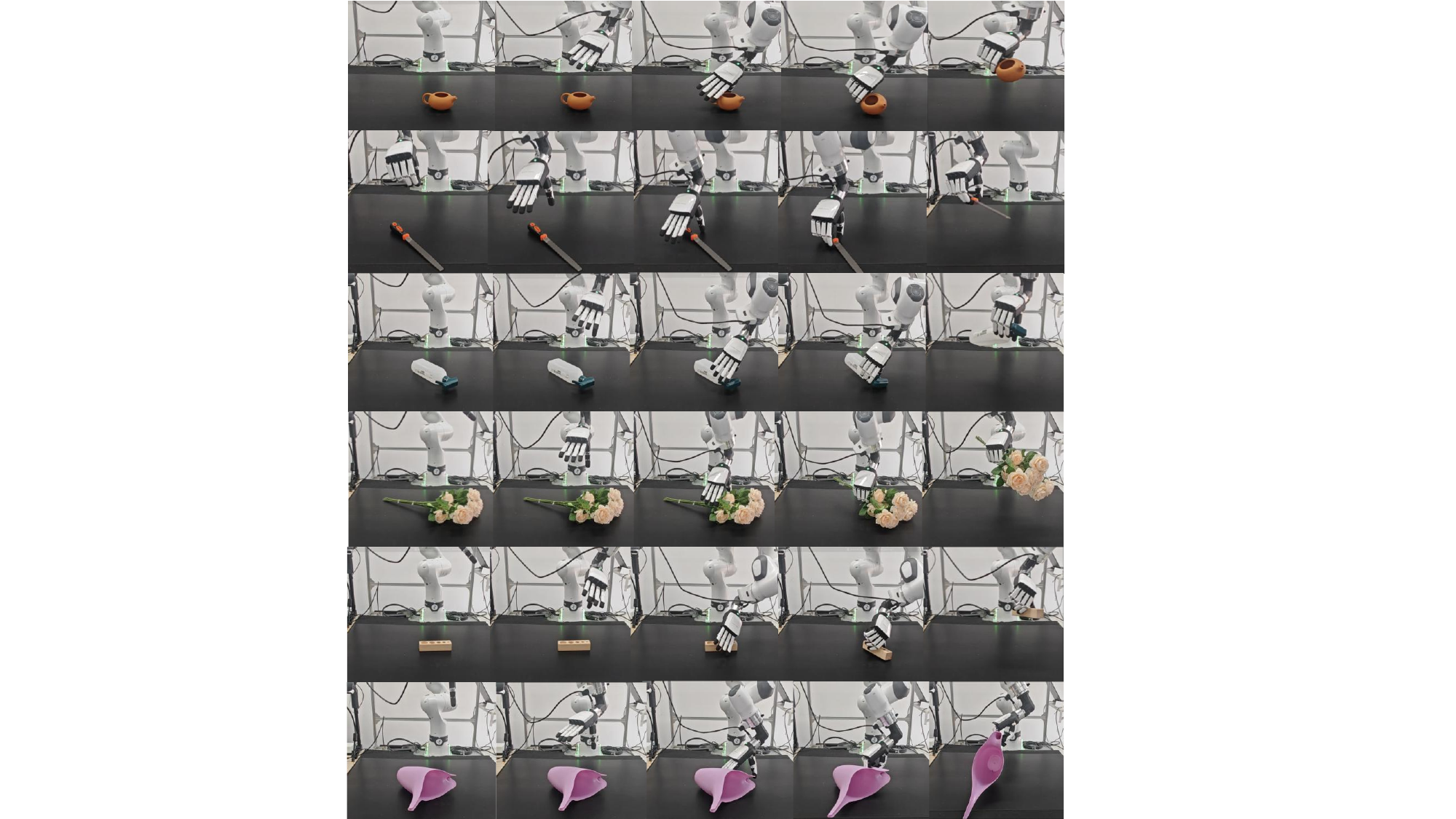}
    \caption{Real-world video recordings of the vision-based policy performing functional grasps on a variety of objects.}
    \label{fig:real_demo}
\end{figure*}

Fig.~\ref{fig:real_demo} presents real-world demonstrations, illustrating the effectiveness of sim-to-real transfer and the robustness of our approach. The policy successfully grasps a wide range of challenging objects, including extremely large objects (e.g., a watering can), delicate items (e.g., a bunch of flowers), and heavy tools (e.g., a long metal instrument). These results emphasize the capability of the vision-based policy to generalize from simulation to real-world tasks while maintaining both precision and functional awareness.

Beyond stable grasping, our method can be extended to enable functional manipulation tasks. For example, it can pour water using a teapot or water plants with a spray bottle, demonstrating that the policy not only handles grasping challenges but also executes downstream functional behaviors.


\newpage
\section{Additional Ablation Studies}
\label{sec:Other_Trials}

In addition to the experiments presented above, we also explore several alternative approaches. Although these methods are ultimately suboptimal, they provide useful insights into the challenges of functional grasping.

\textbf{Sampling-based method.}  
We first attempt to collect data using a sampling-based strategy. However, its data efficiency is extremely low (around 5\%), requiring substantial computation to gather a sufficiently large dataset. Even after more than 10k samples, the policy does not learn meaningful affordance cues. Many object segments are \textit{intrinsically difficult} to grasp without refined end-effector rotations and hand joint positions. As a result, the dataset becomes highly imbalanced: trajectories concentrate on only a few graspable segments per object, and a large portion of samples come from objects that are naturally easier to grasp.

\textbf{Planning-based method.}  
We also train a policy to predict a pre-grasp rotation and translation, and then plan toward the target affordance via linear interpolation. However, this interpolation strategy severely restricts the feasible action space and makes it difficult to reach certain geometric affordance regions. Experiments on the YCB dataset show that the initial success affordance distance is 3.8\,cm. After optimization with a binary success reward, the success rate increases to over 60\%, but the mean success affordance distance increases to over 6\,cm. These results indicate that although training improves binary success, linear interpolation fundamentally limits grasp accuracy, especially for objects with diverse geometries.

\textbf{Sampling styles from successful trajectories.}  
We further analyze successful DemoGrasp trajectories and sample hand-style distributions as priors for style adaptation. However, the resulting hand poses are neither human-like nor stable for policy learning. The ``diverse'' grasp styles lack correlation with object geometry and often lead to loose or suboptimal grasps. This suggests that human-like grasping styles are inherently object-dependent and that geometry-aware style generation is essential for achieving tight, functional grasps.





\clearpage

\end{document}